\documentclass[review]{elsarticle}
\graphicspath{ {./figures/} }
\usepackage{float}
\usepackage{verbatim} 
\usepackage{apalike}
\restylefloat{figure}
\floatstyle{plaintop}
\restylefloat{table}
\usepackage{placeins}
\usepackage{array, makecell}
\usepackage[ruled,vlined]{algorithm2e}

\usepackage[shortlabels]{enumitem}
\setlist[enumerate]{nosep}
\usepackage{listings}
\usepackage{blindtext}
\usepackage{mathtools}
\usepackage{amsmath}

\usepackage{amsfonts}
\usepackage{graphicx}
\usepackage{tabularx}
\usepackage{multirow}
\usepackage{rotating}
\usepackage[colorinlistoftodos]{todonotes}
\usepackage[colorlinks=true, allcolors=blue]{hyperref}
\newcommand{\beq}{\begin{equation}}
\newcommand{\eeq}{\end{equation}}
\newcommand{\bde}{\begin{definition}}
\newcommand{\ede}{\end{definition}}
\newcommand{\bpp}{\begin{property}}
\newcommand{\epp}{\end{property}}
\newcommand{\bpr}{\begin{proposition}}
\newcommand{\epr}{\end{proposition}}
\newcommand{\bex}{\begin{example}}
\newcommand{\eex}{\end{example}}
\newcommand{\bco}{\begin{corollary}}
\newcommand{\eco}{\end{corollary}}
\newcommand{\bre}{\begin{remark}}
\newcommand{\ere}{\end{remark}}
\newcommand{\bal}{\begin{algorithm}}
\newcommand{\eal}{\end{algorithm}}
\DeclareUnicodeCharacter{2212}{-}
\newcommand{\ble}{\begin{lemma}}
\newcommand{\ele}{\end{lemma}}

\journal{Applied Soft Computing}
\definecolor{bluesky}{RGB}{0, 0, 0}
\newcommand{\revised}[1]{{\color{bluesky}{#1}}}

\newcommand{\commentcode}[1]{{\color{cyan}{#1}}}
\biboptions{authoryear}
\usepackage{hyperref}
\begin{document}
\begin{frontmatter}

\begin{titlepage}
\begin{center}
\vspace*{1cm}

\textbf{ \large Multi-Objective Genetic Algorithm for Multi-View Feature Selection}

\vspace{1.5cm}

Vandad Imani$^{a}$ (vandad.imani@uef.fi), Carlos Sevilla-Salcedo$^c$ (carlos.sevillasalcedo@aalto.fi), Elaheh Moradi$^a$ (elaheh.moradi@uef.fi), Vittorio Fortino$^b$ (vittorio.fortino@uef.fi), Jussi Tohka$^a$ (jussi.tohka@uef.fi) \\

\hspace{10pt}

\begin{flushleft}
\small  
$^a$ A. I. Virtanen Institute for Molecular Sciences, University of Eastern Finland, Finland \\
$^b$ Institute of Biomedicine, University of Eastern Finland, Finland \\
$^c$ Department of Computer Science, Aalto University, Espoo, Finland

\vspace{1cm}
\textbf{Corresponding Author:} \\
Vandad Imani \\
Email: vandad.imani@uef.fi

\end{flushleft}        
\end{center}
\end{titlepage}

\title{Multi-Objective Genetic Algorithm for Multi-View Feature Selection}

\author[label1]{Vandad Imani \corref{cor1}}
\ead{vandad.imani@uef.fi}

\author[label3]{Carlos Sevilla-Salcedo}
\ead{carlos.sevillasalcedo@aalto.fi}

\author[label1]{Elaheh Moradi}
\ead{Elaheh.Moradi@uef.fi}

\author[label2]{Vittorio Fortino}
\ead{vittorio.fortino@uef.fi}

\author[label1]{Jussi Tohka}
\ead{jussi.tohka@uef.fi}

\author {for the Alzheimer’s Disease Neuroimaging Initiative \corref{cor2}}

\cortext[cor1]{Corresponding author.}

\address[label1]{A. I. Virtanen Institute for Molecular Sciences, University of Eastern Finland, Finland}
\address[label2]{Institute of Biomedicine, University of Eastern Finland, Finland}
\address[label3]{Department of Computer Science, Aalto University, Espoo, Finland}

\begin{abstract}
Multi-view datasets offer diverse forms of data that can enhance prediction models by providing complementary information. However, the use of multi-view data leads to an increase in high-dimensional data, which poses significant challenges for the prediction models that can lead to poor generalization. Therefore, relevant feature selection from multi-view datasets is important as it not only addresses the poor generalization but also enhances the interpretability of the models. Despite the success of traditional feature selection methods, they have limitations in leveraging intrinsic information across modalities, lacking generalizability, and being tailored to specific classification tasks. We propose a novel genetic algorithm strategy to overcome these limitations of traditional feature selection methods for multi-view data. Our proposed approach, called the multi-view multi-objective feature selection genetic algorithm (MMFS-GA), simultaneously selects the optimal subset of features within a view and between views under a unified framework. The MMFS-GA framework demonstrates superior performance and interpretability for feature selection on multi-view datasets in both binary and multiclass classification tasks. The results of our evaluations on nine benchmark datasets, including synthetic and real data, show improvement over the best baseline methods. This work provides a promising solution for multi-view feature selection and opens up new possibilities for further research in multi-view datasets.
\end{abstract}


\end{frontmatter}

\section{Introduction}
\label{introduction}

Machine learning problems are often defined with features from different views (or modalities), where each "view" (or "modality") represents a particular characteristic of the data. The fusion of multi-view data enhances the performance of machine learning algorithms in solving a particular task, such as classification, by incorporating complementary information from each view. However, a challenge associated with multi-view data fusion is the curse of dimensionality, which can make it difficult to extract valuable knowledge from the massive feature space that is created by combining multiple views, and may contain irrelevant and redundant features. Therefore, the use of conventional machine learning approaches can lead to overfitting, impairing the generalizability of the algorithm, and deteriorating its performance~\citep{pappu2014high}. One solution to this challenge is to use feature selection (FS), which reduces dimensionality by identifying the most relevant subset of features and eliminating redundant features~\citep{ma2021two}.

A simple approach to integrating multi-view data is feature concatenation, which involves combining all available views into a single vector and then applying the FS algorithms to extract the most informative features. However, this strategy introduces three new difficulties for feature selection in the multi-view setting~\citep{xie2011m}. First, the curse of dimensionality limits the performance improvement when the number of features far exceeds the number of training subjects available. Second, it is challenging to explore the complementary information of different feature groups and select the appropriate features for classification, since different features of various views have distinctive roles in the task. Third, the performance of the model will easily deteriorate if any features in the concatenated representations are redundant or irrelevant to the classification task. Therefore, it would be ideal to provide an effective way to evaluate multi-view features and select the relevant and informative.

Due to the aforementioned difficulties, several multi-view FS algorithms have recently been developed, primarily based on filter~\citep{liu1996probabilistic}, embedded~\citep{guyon2003introduction}, and wrapper~\citep{kohavi1997wrappers} models.~\cite{jha2021incorporation} proposed a filter-based multimodal multi-objective optimization using ring-based particle swarm optimization algorithms with a specific crowding distance to evaluate the relevance of a subset of features to its intrinsic properties. This filter-based technique scores features using an evaluation index independent of the classification learning algorithm. However, this is typically less robust, as the FS process is not directed by the classification performance. Further, the selected subset of features may not be the smallest, which might be prohibitive in cases with high number of features~\citep{hu2022dispersed}. 

The embedded methods~\citep{lei2017joint,imani2021comparison,sevilla2022multi} perform FS while learning the classifier and incorporate it either within the algorithm or as an additional functionality. The multitask feature selection techniques presented in ~\citep{zhang2011multimodal,liu2014inter,chen2018multi,liu2020enhancing} are typical examples of embedded approaches, as they choose a shared subset of relevant features from each modality.  
Specifically,~\cite{jie2015manifold} proposed a manifold regularized multitask feature learning method to preserve the complementary inter-modality information by introducing a manifold-based Laplacian regularizer to embed the manifold information in the FS.~\cite{zu2016label} proposed a label-aligned multi-task feature learning method in which a new label-aligned regularization term is added to the objective function of standard multi-task feature selection to find the most discriminative subset of features.~\cite{shi2022asmfs} proposed an adaptive similarity-based multi-modality feature selection method capable of simultaneously capturing the similarity among multi-modality data and enabling joint learning of feature selection in which the similarity matrix and FS are alternately updated. These embedded methods focus on capturing an optimal subset of features from each view, including fully noisy modalities. Consequently, the performance of the classifier will rapidly deteriorate due to the selection of noisy features from the fully noisy modality, making it unsuitable as a unified method for preprocessing big data~\citep{hu2022dispersed}. In addition, these methods require the same number of features from different views, which considerably limits the usability of the model. 

Wrapper-based algorithms~\citep{caruana1994greedy,liu2005toward,ma2017novel,sayed2018novel} aim to identify the most effective subset of features for the target prediction task. This is done through a search process, where different subsets of features are evaluated based on their classification performance. To evaluate a subset of features, the algorithm maps the data to only include the selected features, trains a classification model using this reduced set of features, and then evaluates the model's performance on a validation set. This iterative process continues until the algorithm identifies the subset of features that results in the best classification performance. In multiview settings, sequential forward feature selection (SFFS)~\citep{sheng2019novel,li2019hierarchical} and sequential backward feature selection (SBFS)~\citep{liu2018electroencephalogram} have been used to address the problem of unifying small representations of heterogeneous features from different views. However, they are susceptible to the so-called "nesting effect", which occurs when a feature is discarded using the top-down method and cannot be reincorporated to the selected subset~\citep{yusta2009different}. Consequently, the final optimal subset may not include all desirable features. Biology-based meta-heuristics algorithms such as evolutionary computational approaches are excellent means of overcoming the nesting effect and stalling in local optima since they have no restrictions on selecting features throughout their search process~\citep{yue2019multimodal}. 

Concentrating on the challenges stated above, we propose a novel genetic algorithm (GA) strategy, to simultaneously select the optimal subset of features as well as the optimal subset of views under an enhanced framework of multi-view multi-objective feature selection (MMFS-GA). MMFS-GA selects the optimal features for each view by exploring a comprehensive range of potential feature combinations in population. \revised{This selection is further refined by alternating between single-view feature selection and multi-view combination selection, thereby optimizing the use of complementary and supplementary information across views. This process evaluates features from individual views in the context of their collective impact on model performance, enabling the identification of complementary interactions that might be missed by simple concatenation.}

This population-based method makes the unification of heterogeneous features from different views meaningful because it allows each data view with distinct dimensions and properties to be combined appropriately. We employ a multiniche multi-objective GA, which preserves population diversity, by segmenting the GA population into several semi-disjoint population sets. In addition, unifying features under a multiniche framework is a more principled approach to circumventing the problems of different representations between views. This is because it selects the most stable solutions out of an extensive search region in terms of the classifier's performance. Our principal contributions are summarized as follows:
\begin{enumerate}
  \item[(1)] \revised{We propose MMFS-GA, a novel multi-view feature selection method that combines two multi-objective genetic algorithms based on niching NSGA-II~\citep{deb2002fast}. This allows stable feature selection from each distinct dimensional view while suppressing irrelevant views that behave as classification noise.
  
  \item[(2)]  We further propose a duplicate elimination strategy that fine-tunes the balance between convergence and diversity in the population space, enhancing the exploration of informative features.}
  \item[(3)] The proposed method is evaluated against embedded and wrapper methods in binary and multiclass classification tasks across synthetic and real-world datasets. \revised{In our validation, datasets covered both consistent feature dimensions and those with varied feature counts across views. The experimental results show that our method outperforms competitors in accuracy, feature selection efficiency, and in determining relevant views across classification tasks.}
\end{enumerate}

In Section~\ref{Background}, we describe the MMFS-GA prerequisites. Section~\ref{Method} explains the MMFS-GA algorithm. The experimental results are presented in Section~\ref{results}. In Section~\ref{discussion}, we discuss the proposed method against other relevant methods in the literaure before moving on to Section~\ref{conclusion}, where we draw the conclusions.

\section{Background}
\label{Background}

\subsection{Genetic Algorithms for feature selection}

Genetic algorithms (GAs) are a heuristic solution-search and optimization technique inspired by biological evolution. They employ a simplified form of evolutionary processes to find solutions to complex optimization problems. 
To do so, they are initialized with a randomly generated population of candidate solutions, called chromosomes, which are iteratively modified by genetic operators. To breed new solutions for the next generation, the chromosomes of the population are selected on the basis of their fitness value, which measures the performance of the chromosome on a particular problem. Then, genetic operators, i.e., crossover and mutation, are applied to the selected chromosomes to produce offspring for the new population. This process iterates through a series of generations, during which the fitness value of the chromosomes tends to improve until a stopping criterion is reached.

GAs have been frequently used as wrapper-based techniques to identify the best subset of features. In these methods~\citep{reddy2020hybrid,dong2018novel,zhu2007markov,oh2004hybrid,raymer2000dimensionality}, the chromosome is represented as a binary vector to determine the use of features. For each feature, the value of "1" indicates that it is part of the selected set of features, while a value of "0" indicates that it is not. A classifier evaluates each chromosome in a population by training the model with the selected features and providing the accuracy of the model back to the GA as a fitness value.

\subsection{Multi-objective Optimization Problems}
Multi-objective optimization refers to an optimization problem 
that requires the optimization of two or more conflicting objectives:
\begin{equation}
\label{Eq_1:MOP}
\min_\mathbf{x} f(\mathbf{x}) = (f_{1}(\mathbf{x}),f_{2}(\mathbf{x}),\ldots,f_{m}(\mathbf{x}))
\end{equation}
subject to
\begin{equation} 
\begin{split} 
l_{j}(\mathbf{x}) &\geq 0,\quad  j=1,2,\ldots,J\\
h_{k}(\mathbf{x}) &= 0,\quad    k=1,2,\ldots,K
\end{split}
\end{equation}
where $\mathbf{x} = (x_{1},x_{2},\ldots,x_{v}) $ represents a set of variables in some decision space $\mathcal{D}^v$, $f(\mathbf{x})$ is a set of $m$ objectives to be minimized, and $l_{j}(\mathbf{x})$ and $h_{k}(\mathbf{x})$ are inequality and equality constraints, respectively. 

In contrast to single-objective optimization, it is difficult to determine which solution is the best in multi-objective optimization since the quality of a solution is described in terms of trade-offs between conflicting objectives. A dominating relationship~\citep{konak2006multi} is utilized when comparing the various solutions of multi-objective optimization problems. Let $\mathbf{x} \in \mathcal{D}^v$ be a feasible solution in the decision space, and let $f_i(\mathbf{x})$ denote the value of the i\textsuperscript{th} objective function in $\mathbf{x}$. Then, a solution $\mathbf{x}$ is said to dominate another solution $\mathbf{x}' \in X$  (and we write $\mathbf{x}\prec \mathbf{x}'$) if the following criteria are met:
\begin{equation}
\label{Eq_3:PF}
\begin{cases}
\forall i \in \{1,2,\ldots,m\} : f_{i}(\mathbf{x}) \leq f_{i}(\mathbf{x}') \\[1pt]
\exists j \in \{1,2,\ldots,m\} : f_{j}(\mathbf{x}) < f_{j}(\mathbf{x}')
\end{cases}.
\end{equation}

A good compromise in multi-objective optimization involves identifying a set of solutions that are not dominated by any other feasible solution, meaning that none of these solutions can be improved without negatively impacting at least one of the other objectives. This set of solutions is known as the Pareto-optimal solution set (PS). The Pareto front (PF) is the set of all situations in which no preference criterion can be improved without worsening at least one preference criterion. Non-dominated solutions, also known as Pareto-optimal solutions, are those that belong to the Pareto-optimal solution set and are not dominated by any other feasible solution.

\begin{figure*}[h!]
\centering
\includegraphics[width=0.99\textwidth]{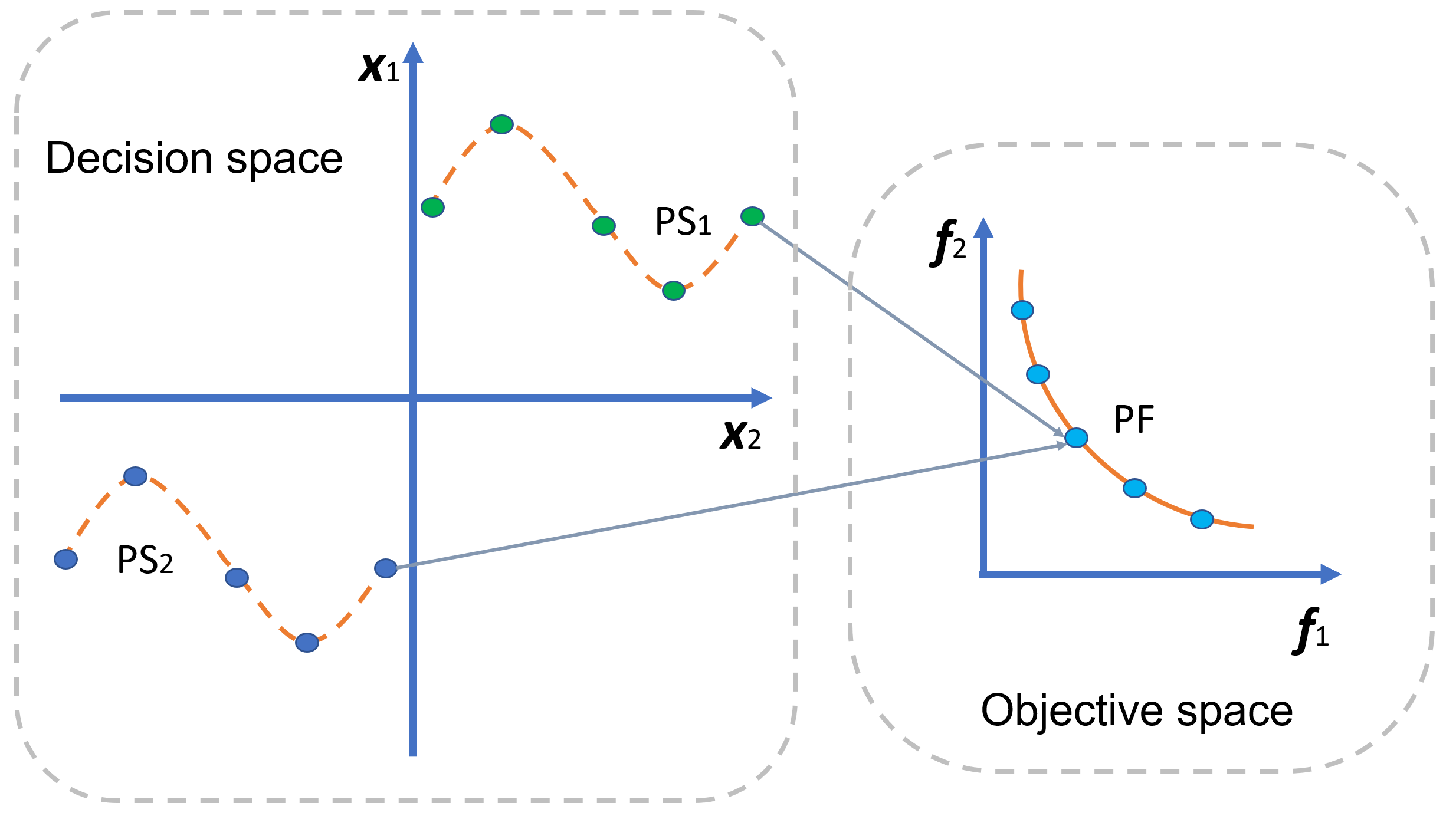}
\caption{An illustrative example of a multimodal multi-objective problem.} 
\label{Fig_1:Multimodal_multiobjective}
\vspace{0mm}
\end{figure*}

\revised{Advantages of a multi-objective approach include a broader exploration of the solution space along the Pareto front, revealing trade-offs between model complexity and performance. This is useful in real-world scenarios where we might need simpler models due to computational limits or where the cost of misclassification is particularly high. By concurrently optimizing two objectives, we mitigate risks such as overfitting or selecting overly complex models, common difficulties encountered in single-objective optimization frameworks~\citep{jensen2004helper}.} In contrast to single-objective optimization, where a unique optimal solution is sought, multi-objective optimization problems often involve finding a set of Pareto-optimal solutions that represent trade-offs between conflicting objectives. The set of all Pareto-optimal solutions in the decision space corresponds to the Pareto front in the objective space. However, multimodal multi-objective optimization problems specifically focus on problems with at least two different Pareto optimal sets in the decision space that corresponding to the same Pareto optimal solution of the PF in the objective space~\citep{liang2016multimodal}. In this context, the term "multimodal" refers to the presence of multiple globally optimal solutions, rather than a multi-view dataset. The multimodal multi-objective optimization problem is depicted in Figure~\ref{Fig_1:Multimodal_multiobjective}, where there are two different solutions that belong to $PS_{1}$ and $PS_{2}$, which correspond to the same PF. A critical challenge with multimodal multi-objective optimization in feature selection is that different feature subsets may have the same objective values, which can lead to ambiguity and hinder the optimization process~\citep{kamyab2016feature}.

\subsection{NSGA-II algorithm for multimodal multiobjective optimization}
\label{NSGA}
Evolutionary algorithms, such as GAs, which are based on the idea of evolving a multiset of candidate solutions and subsequent selection, are commonly employed to solve multimodal multi-objective optimization problems. The Nondominated Sorting Genetic Algorithm II (NSGA-II)~\citep{deb2002fast} is one of the most widely used algorithms to solve multimodal multi-objective optimization problems. It has three unique characteristics that make it stand out from other optimization algorithms: a simple crowded comparison operator, a fast crowded distance estimation procedure, and a fast non-dominated sorting approach. We adapt the selection operator of NSGA-II, which ensures less computational complexity compared to other evolutionary algorithms~\citep{garcia2019comparison} and uses elitism to prevent the loss of optimal solutions throughout iteration~\citep{liang2016multimodal}.

To select a diverse population, NSGA-II first uses a fast non-dominated sorting method to assign a rank to each solution in the population. The solutions in the objective space are sorted into different fronts, $F_{i}$ based on the values of the objective functions. The first front, $F_{1}$, contains solutions that are not dominated by any other solutions in the objective space. $F_{2}$ is comprised of solutions that are dominated by one or more $F_{1}$ solutions, etc. In the case that $i < j$, solutions that come in front $F_{i}$ are better than those that come in front $F_{j}$, as solutions in front $F_{i}$ are non-dominated and closer to the Pareto front than those in front $F_{j}$.

NSGA-II employs a "crowding distance" to maintain the solutions as diverse as possible. The crowding distance measures the density of a solution relative to its neighbors (in the objective space) who lie on the same Pareto-front rank $F$. To calculate the crowding distance, it is necessary to sort the solutions in $F_{i}$ according to their fitness value in order to identify the neighbors of a solution. An infinite crowding distance is assigned to the solutions with the smallest and largest values for each objective function. The distance $d_i$ for intermediate solutions is calculated based on the absolute normalized difference between the fitness values of two neighboring solutions as
\begin{equation}
d_{i} = \sum_{m} \frac{|f_{m}^{i+1} - f_{m}^{i-1}|}{\max f_{m} - \min f_{m}},
\end{equation}
where $f_m^{i - 1}$ and $f_m^{i +1}$ are the two neighboring solutions to $i\textsuperscript{th}$ solution in terms of the objective $f_m$. Equipped with ranking and crowding distance, NSGA-II uses the tournament selection operator to identify good solutions by randomly selecting two solutions and comparing them according to first their ranks and then their crowding distance. Priority is given to the solution with the lowest rank, and if both solutions have the same rank, the one with the largest crowding distance is selected.

NSGA-II has been applied to feature selection problems because of its ability to handle multiple objectives and constraints. For instance,~\cite{hamdani2007multi} proposed a multi-objective feature selection approach based on NSGA-II that aims to simultaneously optimize classification accuracy and feature subset cardinality. \cite{xue2023feature} built upon this work by incorporating the ReliefF algorithm into the NSGA-II framework to further enhance the performance of FS. In both studies, the ranking and crowding distance mechanisms of NSGA-II were used to maintain a diverse and balanced population of solutions and identify the best feature subsets based on their Pareto fronts. However, these methods were developed for single view data. In addition, \cite{cui2020mmco} proposed a multi-view FS method for cluster analysis of high-dimensional gene expression data. To the best of our knowledge, this represents the only instance where NSGA-II has been applied to FS using multi-view data, but no methods building on NSGA-II have been proposed for multi-view supervised classification.   

\section{The proposed method: MMFS-GA}
\label{Method}
In this section, we present the proposed Genetic Algorithm for Multi-view Multi-objective Feature Selection, MMFS-GA. We provide details of the proposed algorithm framework, followed by the optimization algorithm and the classification models employed.

\subsection{Problem Definition}
Let $\mathbf{X} = [\mathbf{X}_1,\mathbf{X}_2,\ldots,\mathbf{X}_V] \in \mathbb{R}^{e \times k}$ be a multi-view dataset including $e$ samples for which data from $V$ views (or modalities) are available, where $\textbf{X}_v = [\mathbf{x}_{1}^{\revised{v}},\mathbf{x}_{2}^{\revised{v}},\ldots,\mathbf{x}_{e}^{\revised{v}}] \in \mathbb{R}^{e \times k_{v}}$ represents the data matrix of $v\textsuperscript{th}$ view (or modality), and ${\mathbf{x}}_{i}^{\revised{v}}\in \mathbb{R}^{1 \times k_{v}}$ is the within-view features of $i\textsuperscript{th}$ sample ($i = 1,2,\ldots,e$). The total dimensionality of the samples is $k = \sum_{v=1}^{V} k_{v}$, where $k_{v}$ is the dimensionality of the $v\textsuperscript{th}$ view. Then, $\textbf{y} = \{y_{1},\ldots, y_{e} \}$ is the set of all response variables, where $y_{i} \in \mathbb{Z}$. In particular, $y_{i} \in \{0, 1\}$ in the binary classification task and $y_{i} \in \{1, 2, \ldots ,C\}$ in the multiclass classification.

We aim to use GA to identify important features in the input data $\mathbf{X}$ via multiniche, multi-chromosome design. Let $ \mathbf{p}\in \{0,1\}^{1 \times k}$ be a set of membership indicators for features in $\textbf{X}$ where $p_j = 1$ denotes the presence of the feature $x_j$ in the optimum set of features and $p_j = 0$ indicates its absence. We aim to minimize the estimated generalization error of the model while keeping the number of features to a minimum. 
To this end, we formulate FS as a multiobjective optimization problem as described in Equation~\ref{Eq_1:MOP}, 
\begin{equation}
\label{Eq_6:ffun}
\begin{split} 
\min_{\mathbf{p}} \quad & \left( f_{1}(\mathbf{p}), f_{2}(\mathbf{p}) \right) \\
\text{where} \quad & f_{1}(\mathbf{p}) =  \mathcal{E}(y,\hat y( \mathbf{X};\mathbf{p})), \\
& f_{2}(\mathbf{p}) =  \sum_{j=1}^{k} p_{j},  \\
\end{split}      
\end{equation}
\text{subject to}
\begin{equation} 
\sum_{j=1}^{k} p_{j} \geq 1,
\end{equation}
where $\mathcal{E}$ refers to the classification error rate (we use a balanced accuracy score) with $\hat y$ representing the estimate of $y$ based on the input variables that are considered to be the most useful ($p_j = 1$).

\subsection{Overview of MMFS-GA Framework}

\begin{figure*}[h!]
\centering
\includegraphics[width=0.99\textwidth]{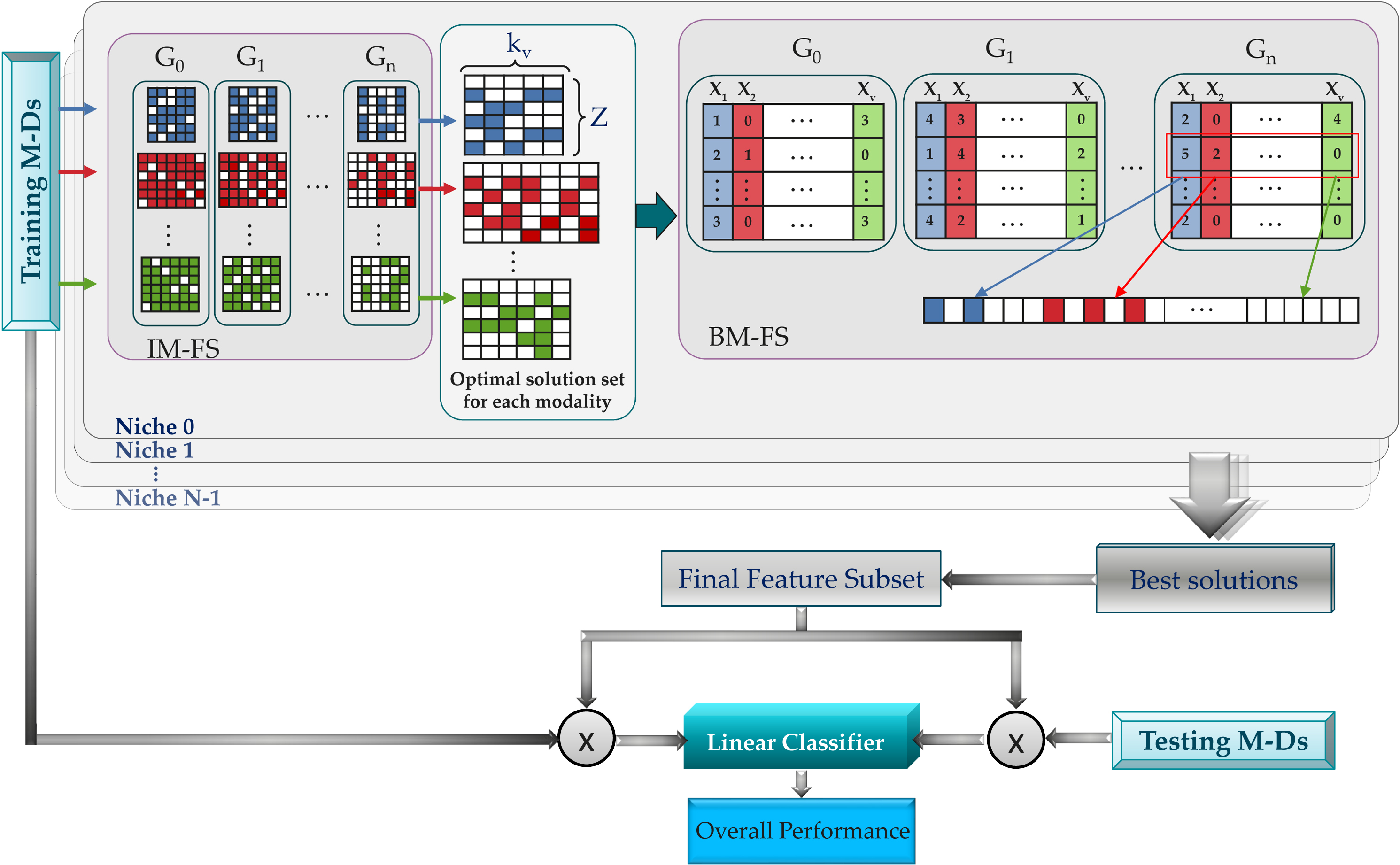}
\caption{The overall framework for multi-view Feature Selection using Genetic Algorithm (MMFS-GA).} 
\label{Fig_2:MMFSGA_framework}
\vspace{0mm}
\end{figure*}

We propose a GA-based framework to solve the feature selection problem posed in Equation~\ref{Eq_6:ffun}. The proposed GA for multi-view feature selection has two main steps: (1) selecting informative features from each view of the data, referred to as intra-view feature selection (IV-FS); and (2) selecting the optimal set of modalities from multi-view data, referred to as between-view feature selection (BV-FS). The pseudocode of MMFS-GA is presented in Algorithm~\ref{alg:MMFSGA}.
To avoid premature convergence, MMFS-GA is a multiniche technique in which $N$ different niches evolve their own populations independently through crossover and mutation. The overall structure of the proposed method is illustrated in Figure~\ref{Fig_2:MMFSGA_framework}.

In the first step, the IV-FS algorithm produces $L = 6$ solutions for each niche ($n \in N$) from each view, denoted $\mathbf{z}_{n,l,v} \in \{0,1\}^{1 \times k_v}$, \revised{$l = 0,\ldots,5;$} encoding the selected features of the $v\textsuperscript{th}$ view. Setting $L =6$ enables a more diverse exploration of feature combinations, thereby increasing the probability of finding optimal solutions in the BV-FS algorithm. The resulting $L$ solutions form a set \revised{$\mathbf{Z}_{n,v}$}, as presented in Algorithm~\ref{alg:MMFSGA}. The sets in \revised{$\mathbf{Z}_{n,v}$} are then used by BV-FS to identify purely noisy views in the dataset and to select the ideal subset of data views. For each niche $n$, the BV-FS algorithm employs integer-encoded chromosomes ($p_{j} \in \{0,1,\ldots,5\}$), where each gene in the chromosome represents the binary-encoded solution in \revised{$\mathbf{Z}_{n,v}$}, in order to incorporate the interaction of features in different views. A set of best solutions $\mathbf{B} = [\mathbf{b}_1,\mathbf{b}_2,\ldots,\mathbf{b}_N] \in \{0,1\}^{ N \times k}$ is derived from all $N$ niches to select discriminative features from the original set of features, where $\mathbf{b}_n \in \{0,1\}^{ 1 \times k}$ denotes the solution corresponding to the $n\textsuperscript{th}$ niche. After evaluating solutions from all niches, MMFS-GA provides a new feature space that includes the most discriminative features and the optimal combination of views. 
\RestyleAlgo{ruled}
\begin{algorithm}[h!]
\scriptsize
\caption{MMFS-GA}\label{alg:MMFSGA}
\SetAlgoLined
\SetKwInput{KwData}{Input}
\KwData{$\textbf{X}$ } 
\SetKwBlock{DoParallel}{do in parallel for $n \in N$ niches}{end}
 \DoParallel{\commentcode{\# Execute \textbf{IV-FS ($\textbf{X}_v$)} function to find the view-specific solutions.}\\
\For{$v=1,2,\ldots,V$ }{
 \nl  $\text{ind},~g_{m} \leftarrow \text{number of features in } \textbf{X}_v,~\text{migration rate}$.\\
 \nl  $\textbf{if } \text{ind} < 100 \textbf{ then } \text{pop} = 100 \textbf{ else } \text{pop} = 200.$\\
 \nl Initialize population ($\mathbf{P}^{bin}_n$) from a binomial distribution with $\text{pop}$ individuals.\\
 \nl Evaluate each individual in the population 
 with the fitness function Eq. (\ref{Eq_6:ffun}).

\For{$g =1$, \ldots,$gen$}{
\nl $\mathbf{R}^{bin}_n$ = CrossoverOrMutation($\textbf{P}^{bin}_n$). \commentcode{\# Algorithm (\ref{alg2:CrossoverORMutation})}

\nl $\mathbf{P}^{bin}_n$ = Selection ($\mathbf{R}^{bin}_n$).\commentcode{\# Section (\ref{NSGA})}

\commentcode{\# If diversity is low, then remove the duplicate individuals (Section (\ref{Duplicate}))}

\nl \If { Similarity ($\mathbf{P}^{bin}_n$) $>$ 0.8}{
$\textbf{P}^{bin}_n$ = CrossoverAndMutation ($\mathbf{P}^{bin}_n$)}
\nl \If { $\mod(g,g_{m}) == 0$}{
\commentcode{\# Migrate the population between niches.}\\
$\textbf{P}^{bin}_{1, \ldots,N}$ = Migrate($\textbf{P}^{bin}_{1, \ldots,N}$).\commentcode{\# Section(\ref{Multiniche}).}}

\commentcode{\# If 30\%, 60\%, or 90\% of generations have passed, store the best solutions based on accuracy.
}

\nl \If { $mod(g,l*gen*0.3) == 0$ for $l \in \{3,4,5\} $}{
$\mathbf{z}_{n,l,v}\leftarrow SelectBest(\mathbf{P}^{bin}_n)$}}

\nl $\mathbf{z}_{n,1,v} \leftarrow SelectBest(\mathbf{P}^{bin}_n$)\\
\nl $\mathbf{z}_{n,2,v} \leftarrow SelectFrequent(\mathbf{P}^{bin}_n$)\\
\nl $\mathbf{Z}_{n,v} = [ \mathbf{z}_{n,0,v}, \mathbf{z}_{n,1,v}, \mathbf{z}_{n,2,v},\mathbf{z}_{n,3,v}, \mathbf{z}_{n,4,v}, \mathbf{z}_{n,5,v}]$. \commentcode{\# Optimal solutions set.}}
\commentcode{\# Execute \textbf{BV-FS ($\mathbf{X}$, $\mathbf{Z}_{n,v}$)} algorithm to find the best between-view solution.}\\

\nl Initialize population ($\mathbf{P}^{int}$) of 50 individuals, with a integer value between 1-size($\mathbf{Z}_{n,v}$).

\nl Evaluate each individual in the population according to the fitness function.

\For{$g =1$, \ldots,$gen$}{

\nl $\mathbf{R}^{int}_n$ = CrossoverOrMutation($\textbf{P}^{int}_n$). \commentcode{\# Algorithm (\ref{alg2:CrossoverORMutation})}

\nl $\mathbf{P}^{int}_n$ = Selection ($\mathbf{R}^{int}_n$).\commentcode{\# Section (\ref{NSGA})}}
\nl $\mathbf{b}_n \leftarrow SelectBest(\mathbf{P}^{int}_n$)\\
}
\nl $\mathbf{B}$ = [$\mathbf{b}_1,\mathbf{b}_2$,\ldots,$\mathbf{b}_N$]\\
\nl $Best_{Chr} \leftarrow BestAccuracy(\mathbf{B})$.\\
\Return $Best_{Chr}$ 
\end{algorithm}

\subsection{IV-FS: multi-view feature selection}
IV-FS attempts to find the most informative features of each view by generating a population of solutions using a parallelized GA. The IV-FS algorithm starts searching for the optimum solution in $N$ parallel niches by initializing a random population $\mathbf{P}^{bin}_n$ of binary vectors whose elements represent the $k_{v}$-dimensional feature search space of the optimization problem. For simplicity, we will remove the niche index ($n$) from the following equations, since the algorithm does not depend on the niches.

We generate a new offspring population $\textbf{Q}$ applying the variation operator to the existing population $\mathbf{P}^{bin}$. The variation operator produces new offspring through crossover or mutation to develop better solutions that will emerge in the population during evolution~\citep{fortin2012deap}. The variation operator is presented in Algorithm ~\ref{alg2:CrossoverORMutation}. \RestyleAlgo{ruled}
\begin{algorithm}
\scriptsize
 \caption{Creating an offspring population via CrossoverOrMutation}\label{alg2:CrossoverORMutation}
\SetKwInput{KwData}{Input}
\KwData{$\textbf{P}, \rho_{\text{crossover}}, \rho_{\text{mutation}}$}    
 \KwResult{The new population $\mathbf{R}$.}
$\textbf{Q}$ = [ ]\\
\While{size($\textbf{P}$) $\neq$ size($\textbf{Q}$)}{
var $r \coloneqq$ a random number between 0 and 1.

\uIf{$r < \rho_{\text{crossover}}$}{
$p_1, p_2$ = Select.random($\textbf{P}$)\\
$q_1, q_2$ = Crossover($p_1$, $p_2$)\\
$\textbf{Q} = \textbf{Q} \cup q_{1}$
}\uElseIf{$r < \rho_{\text{mutation}} + \rho_{\text{crossover}}$}{
$p_1, \sim$ = Select.random($\textbf{P}$)\\
$q_1$ = Mutation($p_1$)\\
$\textbf{Q} = \textbf{Q} \cup q_{1}$
}\Else{
$p_1, \sim$ = Select.random($\textbf{P}$)\\
$\textbf{Q} = \textbf{Q} \cup p_{1}$}}
$\mathbf{R} = \textbf{P}  \cup \textbf{Q}$\\

\textbf{return} $\mathbf{R}$; 
\end{algorithm}
In the case of the crossover process, the elements of two solutions of the parental population $\mathbf{P}^{bin}$ mate to produce a single offspring $\textbf{q}$. The IV-FS algorithm uses binomial crossover to generate the offspring since this method is less dependent on the size of the population~\citep{zaharie2009influence}. In the event of mutation, an element of one solution from the parental population $\mathbf{P}^{bin}$ is selected at random and mutated according to the probability rate of mutation to produce a single offspring $\textbf{q}$. The new population $\textbf{R}$ is created by combining the parental population $\mathbf{P}^{bin}$ with the offspring population $\textbf{Q}$ acquired by applying the variation operator. Note that we employ one operation, crossover or mutation, at a time to produce new offspring.

Prior to selecting individuals for the next generation, $f_{1}$ is computed for each individual in population $\textbf{R}$ by training a suitable classifier and evaluating it via 10-fold cross-validation. As classifiers, we use linear discriminant analysis (LDA) for binary classification tasks and multinomial logistic regression (MLR) for multiclass classification tasks, as described in Sections 2.7 and 2.8 of the supplementary material. This is followed by the calculation of $f_{2}$ for each individual. By the calculation of $f_{1}$ and $f_{2}$ according to Equation~\ref{Eq_6:ffun}, the position of the solutions in the objective space is determined as $S = \{(f_{1,1},f_{2,1}),(f_{1,2},f_{2,2}),\ldots,(f_{1,\mathbf{R}},f_{2,\mathbf{R}})\}$.

The tournament selection operator, adopted by the NSGA-II algorithm as the selection operator (see Section \ref{NSGA}), works by randomly selecting two solutions from the population, comparing the solutions with respect to their front ranks and their crowding distance, and selecting the best one. For example, the better solution between $s_{t} = (f_{1,t},f_{2,t})$ and $s_{u} = (f_{1,u},f_{2,u})$ is determined as follows:
\begin{align}
\begin{aligned} 
\forall t,u \in \{1,2,\ldots,R\}, t \neq u, \qquad\hat S  =& \begin{cases}
          s_{t}  \qquad if \; \; F_{t} < F_{u}\\[1pt]    
          s_{u}  \qquad if \; \; F_{t} > F_{u}\\[1pt]
          s_{t}  \qquad if \; \; F_{t} = F_{u} \; and \; d_{t} > d_{u}\\[1pt]
          s_{u}  \qquad otherwise.
     \end{cases}
\end{aligned}
\end{align}
where $F$ is the front rank and $d$ is the crowding distance of the corresponding solution. In IV-FS, $N$ niches independently evolve their populations through crossover and mutation; nevertheless, niches interact with each other every $5\%$ of the total generations through a genetic operator termed migration, which swaps the top 25\% of their populations.

For each view, the IV-FS algorithm outputs a set of optimal solutions $\revised{\mathbf{Z}_{v}} = \{ \mathbf{z}_{0,v}, \mathbf{z}_{1,v}, \mathbf{z}_{2,v},\mathbf{z}_{3,v}, \mathbf{z}_{4,v}, \mathbf{z}_{5,v} \}$, where $\mathbf{z}_{0,v}$ is a $k_{v}$-dimensional zero vector containing all zeros\revised{, indicating that no features are selected from the view $v$. The inclusion of $\mathbf{z}_{0,v}$ enables our algorithm to thoroughly evaluate all possible feature subsets, including the null set, for each view. This is important for achieving an unbiased and comprehensive multi-view feature selection. Each of the other vectors $\mathbf{z}_{l,v}$, for $l \in \{1,2,3,4,5\}$, is a $k_{v}$-dimensional binary vector that reflects the best set of features during the evolution of IV-FS.}  
Specifically, $\mathbf{z}_{1,v}$  is the solution with the best $f_1$ after $gen$ generations. $\mathbf{z}_{2,v}$ is a solution that incorporates features selected based on their frequency of occurrence within the final population. To determine which features are included in $\mathbf{z}_{2,v}$, we calculate the selection frequency $a_j$ for each feature $j$ within the final population after $gen$ generations. This selection frequency represents the ratio of individuals in the population that have selected feature $j$ to the total number of individuals. If the selection frequency $a_j$ is greater than 0.5, indicating that more than half of the individuals have selected feature $j$, then feature $j$ is included in $\mathbf{z}_{2,v}$.
This ensures that the selected features are representative of a majority of the population. The rest of the solutions ($\mathbf{z}_{3,v}$, $\mathbf{z}_{4,v}$, and $\mathbf{z}_{5,v}$) are identified during intermediate generations based on the best cross-validated $f_1$. 

\subsection{BV-FS: multi-view feature selection}
BV-FS uses the solutions provided by IM-FS to select views that are important for the classification task and unify features from different views. This strategy highlights the multi-view aspect of the feature selection, as a few potential intra-view feature sets are unified by a second algorithm (BV-FS) to select the best views and view-specific feature subsets for the combined classification model.  Note that BV-FS works separately in $N$ niches and the niche $n$ of BV-FS takes its input from the niche $n$ of IV-FS, i.e., we do not break the niches between IV-FS and BV-FS.

BV-FS creates an initial population of $\mathbf{P}^{int}$ with an integer encoding, with each gene taking values from the range 0 to 5, inclusive. Each gene $p_{j}$ of an individual represents a view and its integer value indexes the solution in $\revised{\mathbf{Z}_v}$. This implies that the number of genes in an individual $\mathbf{p}$ equals the number of views. For instance, $p_{1} = 2$ indicates that the solution $\mathbf{z}_{2,1}$ is considered for view $1$. Note that the if $p_v = 0$ classification model does not consider the features of view $v$. 

The BV-FS algorithm utilizes the same fitness value as IV-FS (see Equation~\ref{Eq_6:ffun}), where $f_{1}$ is the number of selected features in the selected views and $f_{2}$ refers to the classification error rate obtained in the validation set by utilizing the desired features of the selected views. Similarly to IV-FS, a variation operator, such as a two-point crossover or Shuffle mutation, is used to produce a new individual. Individuals with a higher fitness value have a better probability of mating and yielding more "optimal" individuals. In this way, it continues to generate ever-better individuals until the end of the generation. 
\revised{At the end of generation, for each niche, a solution is selected according to the highest $f_{1}$ value from their respective niche's population. The solution from each niche is then inducted into the elite set $\mathbf{B}$. Subsequently, to identify the most promising solution from $\mathbf{B}$, a 10-fold cross-validation is employed within the training set. This validation approach allocates a larger subset of data for validation purposes rather than for training. This methodology aims to enhance the robustness of the validation process, mitigate overfitting, and strengthen the generalizability of our model, which is particularly crucial given dataset size limitations. The solution from $\mathbf{B}$ that exhibits the highest accuracy during this validation phase is then selected as an optimal solution. Note that to evaluate the final solution, we always use a completely external test set (unseen data) or, with synthetic data, we evaluate the analytical generalization performance by computing the true error rate assuming an infinite number of test samples~\citep{raudys1991small}.}

\subsection{Counteracting stagnation}
\label{Duplicate}
An issue with the proposed feature selection approach is that individuals with a high value of $f_{2}$ can constitute the majority of the population and become solutions that are not as high-quality as expected. Duplicate individuals may cause evolution to halt because every individual in the population are very similar, resulting in low diversity and premature convergence. We monitor the similarity between individuals in a population for each generation to avoid this. We use the Jaccard similarity coefficient~\citep{jaccard1912distribution} to identify pairwise similarity between individuals in the population. The Jaccard similarity coefficient is calculated as
\begin{equation}
Similarity = \frac{1}{|\mathbf{P}|} \sum_{i = 1,j > i}^{\mathbf{|P|}} J(\mathbf{p}_{i},\mathbf{p}_{j}). 
\end{equation}
where \revised{$J(\mathbf{p}_{i},\mathbf{p}_{j}) =  (\mathbf{p}_{i} \cdot \mathbf{p}_{j}) / ((\|\mathbf{p}_{i}\|^{2} + \|\mathbf{p}_{j}\|^{2} ) - \mathbf{p}_{i} \cdot \mathbf{p}_{j})$ is the Jaccard similarity coefficient between two individuals $\mathbf{p}_{i}$ and $\mathbf{p}_{j}$.} 
When individuals in a population have a high level of similarity, the individuals are repaired by applying both crossover and mutation operators with a probability of 0.9. Consequently, this leads to the replacement of duplicate individuals in the population by newly generated offspring. 

\subsection{Multiniche Genetic Algorithm}
\label{Multiniche}
It is essential to maintain good diversity in the decision space to solve multimodal multi-objective optimization problems. The multiniche genetic algorithm (MN-GA) is an extension of GA that aims to overcome the problem of premature convergence, which occurs when the GA reaches a local optimum before finding the global optimum, by maintaining multiple subpopulations, or "niches," instead of a single population, within a unified optimization process~\citep{cedeno1994multiniche}. Each niche evolves independently using the same genetic operators based on different characteristics of the solutions. The use of multiple niches enables the algorithm to simultaneously explore different regions of the solution space, increasing the chance of finding the global optimum rather than getting stuck in a local optimum. Here, we evolve $N$ niches separately, selecting the best feature subset between the niches only at the end of the algorithm.

\begin{figure*}[h!]
\centering
\includegraphics[width=0.99\textwidth]{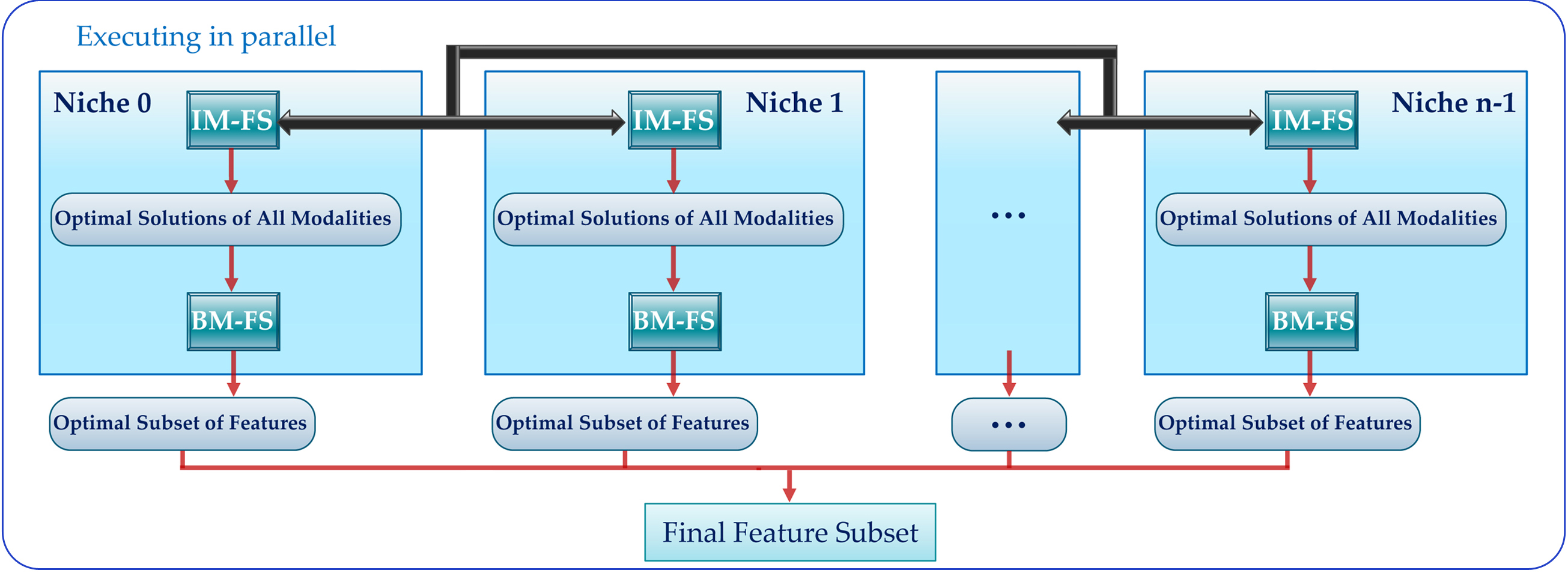}
\caption{The parallel framework for the multiniche Multi-view Multi-objective Feature Selection Genetic Algorithm (MMFS-GA).} 
\label{Fig_3:multiniche_framework}
\vspace{0mm}
\end{figure*}

In MMFS-GA, the migration operator is also used to exchange solutions between different niches, where solutions from one subpopulation are transferred to another subpopulation. This allows for the exchange of genetic information between niches, which increases the diversity of the population and allows for the exploration of different regions of the solution space. Figure~\ref{Fig_3:multiniche_framework} provides a detailed illustration of these processes in the multiniche technique for the MMFS-GA algorithm. The figure highlights two main functions of the MMFS-GA algorithm in each niche. The first function, IV-FS, performs the migration operator in addition to the genetic algorithm operators, while the BV-FS function performs the process of evolution only through crossover, mutation, and selection. The number of niches and the frequency of migration can be adjusted to control the balance between exploration and exploitation in the search process.

\subsection{Implementation}

We implemented the MMFS-GA algorithm using the distributed evolutionary algorithms in Python (DEAP) library~\citep{fortin2012deap}, with a multi-niche schema consisting of six niches, each independently evolving their own populations through crossover and mutation. The niches interacted every 50 generations via a migration genetic operator, which swapped the top 25\% of the best individuals in each niche's population. \revised{The rationale for limiting migration to the top 25\% of individuals is to preserve niche-specific adaptations that have been developed over generations, as detailed in Section 2.5 (Migration size) of the supplementary materials~\citep{den2013maintaining,skolicki2005influence}.} The niches interacted using a queue structure. Each niche contained two functions: IV-FS and BV-FS. In IV-FS, we adjusted the generation and population size for each data view based on the number of features in the view, enabling efficient exploration of the high-dimensional feature space. If the number of features was less than 100, we set the population size and the number generations to 100 and 500, respectively. However, for views with over 100 features, we scaled up the population size to 200 and the number of generations to 1000. In BV-FS, we adjusted the generation size and population size based on the number of views. Specifically, when the number of views exceeded five, we set the parameters for individual and generation to 100 and 600, respectively; otherwise, we halved both parameters. The probability of mating was set at 0.2 for IV-FS and 0.5 for BV-FS, with a mutation probability of 0.1 for both. To address duplicate solutions in IV-FS, we set the probability of mating and mutating solutions at 0.9. \revised{These values were empirically established via preliminary experiments covering feature dimensionalities from 300 to 5000. We consistently applied the same parameters across all experiments and datasets to maintain result consistency and comparability. For a detailed overview of tunable parameters, see the table in supplementary materials, Section 2.6.}  
\revised{We employed two linear classifiers, LDA for binary classification and MLR for multiclass classification, emphasizing our preference for linear models due to their simplicity, interpretability, and computational efficiency. Both classifiers are based on robust statistical theory, with LDA chosen for its effectiveness in distinguishing between two classes in binary scenarios and MLR for its capacity to manage multiclass problems by extending the logistic regression framework. 
While the choice of classifier can impact model performance, we discovered that the performance was more dependent on the feature selection process than on the classifier itself. Both LDA and MLR, which are predicated on a weighted combination of input features for predictions, were considered suitable for our tasks. Our feature selection method aims to identify features conducive to linear decision boundaries, affirming the appropriateness of LDA and MLR for our classification objectives.}  
We implemented both LDA~\footnote{The LDA function is available in the Discriminant Analysis library of scikit-learn, which can be accessed at \url{https://scikit-learn.org/stable/modules/generated/sklearn.discriminant_analysis.LinearDiscriminantAnalysis.html}.} and MLR~\footnote{The "LogisticRegression" function is available in the linear model library of scikit-learn, which can be found at \url{https://scikit-learn.org/stable/modules/generated/sklearn.linear_model.LogisticRegression.html}.} classifiers using publicly available Python libraries. To parallelize the MMFS-GA algorithm, we utilized the functions provided by the Python multiprocessing~\footnote{The multiprocessing module can be found in the official Python documentation at \url{https://docs.python.org/3/library/multiprocessing.html}.} module. Specifically, we used the Process class to create and manage processes, where each process created by the Process class can occupy one core of the CPU. Since we defined six niches, we used only six cores of the CPU.

\section{Experiments and results}
\label{results}
In this section, we present the experimental results, which analyze the performance of the proposed method on two classification datasets, a binary and a multi-class classification dataset, to compare the proposal with various state-of-the-art FS algorithms.

\subsection{Data Description}

\subsubsection{Multi-view Synthetic Dataset}
\label{Multi-view Synthetic Dataset}
We synthesized 2-class and multiclass multi-view classification problems extending the single-view synthesis framework in~\cite{kvrivzek2007improving}. The multi-view datasets consist of five views (or feature groups) each, where two of the feature groups have discriminative power, and the other three groups are different types of noise that are added to evaluate the effectiveness of the proposed approach for filtering irrelevant feature groups. The data is structured in such a way that it is challenging for feature-ranking or greedy feature selection methods to find an optimal feature subset. 

\revised{The detailed configuration of the dataset, including the statistical properties of each view, specific feature settings for binary and multi-class classification, and noise addition strategies, are comprehensively detailed in Section 1.1 of the supplementary material. All feature groups are summarized in Table~\ref{Tab_1:Synthetic_Data_Info}.}
\begin{table}[!ht]
\begin{center}
\setlength{\tabcolsep}{14.0pt} 
\renewcommand{\arraystretch}{1.1}
\caption{Feature groups of the multi-view synthetic dataset. Six and seven informative features of views 1 and 2, respectively, are referred to as $V_{\mathcal{A}} $ and $V_{\mathcal{B}} $.} 
\label{Tab_1:Synthetic_Data_Info}
\resizebox{\textwidth}{!}{
\begin{tabular}{c c c c}
\hline
\hline
\multirow{3}{*}{Feature Group} & \multirow{3}{*}{Characterization} &  \multicolumn{2}{c}{Dimensionality} \\ \cline{3-4} 
              &  &  \makecell{Training set \\ (Binary)} & \makecell{Training set \\ (4-class)} \\ 
 \hline

$View \: 1$   & $V^{\mathcal{A}} $ + Random Gaussian noise with distribution of $ N(0,1) $ & $200 \times 500$ & $400 \times 500$ \\
$View \: 2$   & $V^{\mathcal{B}} $ + Random Gaussian noise with distribution of $ N(0,1) $ & $200 \times 500$ & $400 \times 500$  \\
$View \: 3$   & Random uniform noise with distribution of $ U(0,1) $                       & $200 \times 500$ & $400 \times 500$ \\
$View \: 4$   & Random Chi square noise with distribution of $ \chi^{2}(1) $               & $200 \times 500$ & $400 \times 500$ \\
$View \: 5$   & Random Gaussian noise with distribution of $ N(0,1) $                      & $200 \times 500$ & $400 \times 500$ \\
\hline
\hline
\end{tabular}
}
\end{center}
\end{table}

We generated 10 different multi-view datasets for each classification task with the above procedure, each containing 100 samples per class. For binary classification, we separately estimated the Bayes error rate for views A and B, obtaining values of 0.046 and 0.141, respectively. When the informative views were concatenated, the estimated Bayes error rate decreased to 0.023. For the 4-class classification, we again estimated the Bayes error rate separately for views A and B, obtaining values of 0.069 and 0.193, respectively. When the informative views were concatenated, the optimal Bayes error rate decreased to 0.032. These Bayes error rates provide a benchmark against which to compare the performance of our proposed method.

\subsubsection{TADPOLE Noisy Features Dataset}
\label{Tadpole}

Data were obtained from the Alzheimer’s Disease Neuroimaging Initiative (ADNI) database (adni.loni.usc.edu). Specifically, we used a dataset prepared for the TADPOLE grand challenge~\citep{marinescu2018tadpole} to address the problem of future predictions of AD disease markers and clinical diagnosis using temporal data (\url{https://tadpole.grand-challenge}). We used multi-view data from participants, all corresponding to the first $24$ months, to predict the clinical diagnosis at month 36. The data includes $1,739$ training participants from the initial database, which corresponds to a period of four time steps (baseline, $6$, $12$, and $18$ months). Then, we evaluate our approach in a cohort of $814$ subjects to predict the future clinical diagnosis at month $36$ using data from previous visits of each subject ($6$, $12$, $18$, and $24$ months).
Based on the TADPOLE dataset, we selected six sets of features that are especially relevant for the characterization of Alzheimer's disease; following our previous work \citep{sevilla2022multi}, where a more detailed description of the features can be found. \revised{A summary of the feature groups, including descriptions and the noise augmentation approach, can be found in Table 1 in Section 1.2 of the supplementary material.}

\revised{\subsubsection{TCGA Dataset}
\label{TCGA}
We utilized multimodal data from six cancer types within the Cancer Genome Atlas (TCGA) program (portal.gdc.cancer.gov). These datasets included invasive breast carcinoma (BRCA), brain lower grade glioma (LGG), prostate adenocarcinoma (PRAD), renal cell carcinoma (RCC), ovarian carcinoma (OVCA), and thyroid carcinoma (THCA), encompassing a total of 2,875 subjects across all datasets. For BRCA, the data modalities include methylation data, mRNA expression levels, and copy number variations. Whereas for the other datasets (LGG, PRAD, RCC, OVCA, and THCA), the data modalities consist of methylation data, mRNA, and miRNA expression data. The comprehensive genomic and epigenomic data provided by TCGA enable the detailed identification of cancer subtypes, each defined by unique genomic signatures. These signatures include mutations, DNA methylation patterns, and gene expression profiles. In the TCGA datasets, subtypes are identified by analyzing a range of genomic data, and each subtype is characterized by its distinct genomic signature. This signature may include particular mutations, DNA methylation patterns, or specific gene expression profiles.

Within the BRCA dataset, which includes 734 subjects, the subtypes are Luminal A, Luminal B, HER2-overexpression, basal-like, and normal-like~\citep{dai2015breast}. The OVCA dataset comprises 286 subjects and categorizes subtypes into Differentiated, Immunoreactive, Mesenchymal, and Proliferative~\citep{cancer2011integrated}. In the LGG dataset, including 509 subjects, the subtypes include IDH wild type, IDH mutant without 1p/19q codeletion, and IDH mutant with 1p/19q codeletion~\citep{lv2021effects}. For the PRAD dataset with 330 subjects, the subtypes are classified into ERG fusion and non-ERG fusion~\citep{wang2017significance}. The THCA dataset consists of 493 subjects with BRAF-like and RAS-like subtypes~\citep{lim2023different}. Finally, the RCC dataset, with 523 subjects, includes KICH (renal chromophobe cell carcinoma), KIRC (renal clear cell carcinoma), and KIRP (renal papillary cell carcinoma) subtypes~\citep{hu2019gene}. The demographic details of the subjects in each dataset are outlined in Table 2 in Section 1.3 of the supplementary material.

In binary classification, the objective is to differentiate between two distinct cancer subtypes. Specifically, we compared Luminal A vs. Basal-like subtypes in BRCA, IDH mutant without 1p/19q codeletion vs. IDH mutant with 1p/19q codeletion in LGG, ERG fusion and non-ERG fusion in PRAD, and BRAF-like vs. RAS-like in THCA. These tasks are generally more straightforward due to the often distinct and pronounced genomic disparities between subtypes. For instance, one subtype might exhibit a particular mutation or methylation pattern that is absent in the other, making it easier for classifiers to distinguish between them based on these distinct genomic markers. Conversely, the challenges escalate in multiclass classification, applied to BRCA, LGG, RCC, and OVCA datasets, where the goal is to differentiate among multiple cancer subtypes. Here, the complexity is combined with the more subtle differences and potential overlaps in genomic features across subtypes. Despite each subtype possessing unique genomic traits, the presence of shared characteristics among several subtypes can make it hard to make a clear classification. This overlap can make accurate classification more difficult. Therefore, multiclass classification requires a more detailed and careful approach to effectively differentiate among multiple subtypes.
}

\subsubsection{MCI-to-dementia Conversion Dataset}
\label{Conversion}
One of the most studied questions in machine learning methods applied to clinical neuroscience is to try to predict which people suffering from mild cognitive impairment (MCI) will convert to dementia in a given period of time \citep{ansart2021predicting}. This is an important clinical problem because not all people suffering from MCI convert to dementia, potential interventions are most effective when started early enough, and dementia is a significant public health concern with nearly 10 million new cases each year. It has been demonstrated that this prediction task can be well solved by combining MRI and cognitive testing data at the baseline if 1) the prediction period is relatively short (under 3 years) and 2) the MCI population is at a significant risk of conversion \citep{moradi2015machine}. As our real-world example, we relaxed these assumptions and studied 5 year conversion using MRI and cognitive testing in a more heterogeneous population of MCI consisting of early and late MCIs \citep{edmonds2019early}. 

Similarly to the previous experiment, our data were obtained from the Alzheimer’s Disease Neuroimaging Initiative (ADNI) database (adni.loni.usc.edu), where we extracted participants with MCI status at the baseline (either early or late MCI) and at least 5 years of follow-up. We say that they are converters if their final diagnosis during the follow-up is dementia and non-converters if their final diagnosis is cognitively normal or MCI. This is a considerably more difficult problem than the one studied in, e.g., \citep{moradi2015machine}.  This setting resulted in 332 participants with MRI and cognitive testing data at baseline, of whom 140 converted to dementia. We divided these participants into training and test sets, both with 166 participants. The train and test sets were stratified, i.e., contained an equal proportion of converters and non-converters with the baseline status of late and early MCI. The subject RIDs are available as supplementary material. 

Our data consisted of views: 1) features extracted from T1-weighted MRI (altogether 314 features) and 2) demographic (age, sex, years of education) and cognitive features (altogether 17 features ). As cognitive features, we used all the cognitive test results available in ADNIMERGE table and across distinct phases of the ADNI project. As MRI features, we used the features extracted by the FreeSurfer pipeline \citep{fischl2012freesurfer}. We explain the used features in more detail in the supplementary material as their exact interpretation is not essential to the current technical work.      

\subsection{Baseline Methods}

We apply our MMFS-GA algorithm to the three datasets mentioned above and compare the proposed method with several current state-of-the-art competing methods, including embedded and wrapper methods, to evaluate their performance in selecting informative features. 
\revised{The selected baseline methods were chosen due to their proven effectiveness in multi-view contexts, as evidenced by their application and evaluation in different studies~\citep{haq2008audio,wu2013realistic,shi2022asmfs}.} The baseline methods for multi-view feature selection can be summarized as follows:
\begin{itemize}
    \item Adaptive-similarity-based multi-modality feature selection~\citep{shi2022asmfs} (denoted as ASMFS): An adaptive learning strategy is used to learn similarity measures from multi-modality data and embed this into a multi-modality feature selection framework where $\ell_{1,2}$-norm is employed as a regularization term to learn sparse representations across modalities. The selected features are eventually taken into a multi-kernel support vector machine (Mk-SVM) for classification.

    \item Multi-kernel method with manifold regularized multitask feature learning~\citep{jie2015manifold} (denoted as M2TFS): It uses group-sparsity regularization to jointly select features across different modalities, as well as manifold regularization, which retains the data distribution information by embedding the manifold information into the feature selection algorithm using a predefined similarity matrix. Then, the Mk-SVM approach is utilized to classify multi-view data.
    
    \item Multi-task learning-based feature selection by preserving inter-modality relationships~\citep{liu2014inter} (denoted as IMTFS): It uses group-sparsity regularization to jointly select features across different modalities, as well as impose a constraint to preserve the inter-modality intrinsic relationship among different modalities, with the assumption that these modalities are related to each other. Eventually, the selected features are sent to a Mk-SVM classifier.
    
    \item Multi-kernel method with Sparse Structure-Regularized Learning~\citep{huang2011identifying} (denoted as Lasso-MkSVM): It employs the $\ell_{1}$-norm regularization term to independently learn sparse representations for each modality. The selected features are then classified using an Mk-SVM.
    
    \item LASSO-based~\citep{tibshirani1996regression} feature selection with SVM method (denoted as Lasso-SVM): It employs LASSO for feature selection and then the SVM with a linear kernel for classification.
    
    \item Sequential Forward Floating~\citep{pudil1994floating} Search algorithm (denoted as SFFS): A sequential search algorithm that dynamically changes the number of features by adding or removing features from a candidate subset while evaluating the criterion.

\end{itemize}
 
\revised{The original feature groups were concatenated for the Lasso-SVM and SFFS baseline algorithms to select features and classify the test set using these selected features. In the SFFS method, LDA and MLR were employed not only for evaluating the final classification accuracy for binary and multiclass problems, respectively, but also took on the role of fundamental classifiers within the feature selection process itself. For the other baseline methods, classifiers were chosen as per their original studies to provide a valid comparison under proven conditions.} The implementation of the baseline methods was carried out in MATLAB. The optimal values of hyper-parameters were selected in the 10-fold inner CV loop by maximizing the accuracy score. Since fine-tuning all parameters with a grid search was impractical for M2TFS, IMTFS, Lasso-MkSVM, and Lasso-SVM, we only considered the regularization parameter for controlling the sparsity among all tasks and an optional regularization parameter that controls the norm penalty as the most crucial parameters for the grid search. Both parameters were selected from the candidate set $\{0.001,0.01,0.1,0.2,0.4,0.6,0.8,0.9,1,5,10,30,40,60,100\}$ by maximizing the accuracy score. In the ASMFS method, the sparsity regularization coefficient, the adaptive similarity learning regularization coefficient, and the number of neighbors were selected in the range of $\{0.001,\allowbreak0.01,\allowbreak0.1,\allowbreak1,5,\allowbreak10,\allowbreak30,\allowbreak40,\allowbreak60,\allowbreak100\}$, $\{0.001,0.01,0.1,1,5,10,30,40,60,100\}$ and $\{1,2,\ldots,10\}$, respectively.

\subsection{Performance evaluation} 
With synthetic data, we can compute the actual probability of misclassification (PMC) (assuming an infinite number of test samples) and use conditional PMC as a performance measure. The conditional PMC of a classifier is defined as the PMC of the classifier trained on a given training sample of size $E$ \cite{raudys1991small}. This was approximated using Monte Carlo integration with 10 million simulated test samples. In addition to classification performance, we evaluated the selected features using the F1 score
\begin{equation}
F1 = \frac{TP}{TP + \frac{1}{2}(FP + FN)}, \nonumber
\end{equation}
where $TP$ is the number of selected features known to be informative, $FP$ is the number of selected noise features, and $FN$ is the number of truly informative features not selected. 

\revised{For the TADPOLE and ADNI datasets, our approach was designed to replicate real-world clinical prediction scenarios by splitting the data into different training and test sets. In TADPOLE, 1,739 subjects formed the training set over baseline to 18-month intervals, while an independent set of 814 subjects served for testing 36-month outcome predictions. In the case of the ADNI dataset, 332 participants were split evenly into training and test sets, with each set containing an equal proportion of converters and non-converters, stratified by their baseline status of late and early MCI. To evaluate the performance of MMFS-GA across TCGA datasets, we employed 10-fold cross-validation for each experiment, ensuring robust assessment of predictive accuracy for various cancer types.} 
we computed the balanced accuracy, sensitivity (SEN), specificity (SPE), and area under the curve (AUC) for binary classification tasks and the balanced accuracy, true positive fractions (TPF), and AUC for multiclass classification tasks as external validation metrics in the test set; see Table \ref{Tab_3:Evaluation_metrics}. The AUC for the multiclass classification task was calculated using the one-vs-all approach~\citep{domingos2000well}, where the AUC for each class is calculated by comparing that class against all the other classes combined. 

\begin{table*}[!ht]
\footnotesize
\begin{center}
\caption{For both binary classification tasks, TP represents correctly diagnosed patients with the target condition (AD or MCI), FN represents incorrectly diagnosed patients with the target condition, TN represents correctly diagnosed patients without the target condition (NC), and FP represents incorrectly diagnosed patients without the target condition. FPF refers to false positive fraction.} 
\setlength{\tabcolsep}{20.0pt} 
\renewcommand{\arraystretch}{1.1} 
\resizebox{\textwidth}{!}{\begin{tabular}{l c} 
\hline
\hline
Metric & Classification Tasks \\ \cline{1-2}
 \hline

 $\text{Balanced  accuracy}$ =  $\frac{1}{C} \sum_{c=1}^{C} \frac{TP_{c}}{TP_{c} + FN_{c}}$  & Binary/Multiclass  \\
$\text{Sensitivity} = \frac{TP}{TP + FN}$  & Binary  \\
$\text{Specificity} = \frac{TN}{TN + FP}$ & Binary  \\
$\text{AUC} = \int_{0}^{1} TPF \,d(FPF)$ & Binary  \\
$\text{TPF}_{c} = \frac{TP_{c}}{TP_{c} + FN_{c}}$ & Multiclass  \\
$\text{AUC} = \frac{1}{N} \sum_{c} (N_{c} \times AUC_{c})$ & Multiclass  \\
\hline

\hline
\end{tabular}}
\label{Tab_3:Evaluation_metrics}
\end{center}
\end{table*}
\vspace{0mm}
\subsection{Binary Classification Results}
\subsubsection{Synthetic Dataset}
\revised{This section assesses the MMFS-GA algorithm's performance on uniform-dimension synthetic datasets, providing a controlled setting to isolate and evaluate its feature selection efficacy.}  \revised{We evaluate the actual probability of correct classification (PCC) under the assumption of an infinite number of test samples by using conditional PCC as a performance measure. The conditional PCC was computed using Monte Carlo integration, involving 10 million simulated test samples, in all experiments featuring synthetic data.} The bold numbers denote the best accuracy of the compared methods. Moreover, in supplementary material Section 2.4, we provide a detailed analysis of the computational time of the training phase of our proposed MMFS-GA method. Table~\ref{Tab_4:Binary_BalAcc_SynData} shows the classification performance of all competing methods in the binary classification task\revised{, including scenarios with and without repeated informative views (labelled as 'Synthetic Data' and 'Synthetic Data (rep)' respectively)}.

As seen in Table~\ref{Tab_4:Binary_BalAcc_SynData}, our proposed method outperformed the five other feature selection methods in both standard and repeated view scenarios. This implies that the new feature representation attained by our proposed algorithm can enhance the classification performance for binary tasks, even in the presence of feature redundancy. Specifically, MMFS-GA obtained an estimated overall conditional error rate (CER) across all test sets of 4.3\% in the ten datasets \revised{evaluated without repetition, and a marginally higher CER of 5\% in the datasets with repeated views. This slight increase in error rate in the presence of feature redundancy demonstrates the robustness of MMFS-GA to handle overlapping and redundant features effectively.} Additionally, MMFS-GA was more consistent than competing approaches, with lower standard deviations of balanced accuracy across the ten datasets. These results indicate the value of the proposed method in enhancing the performance and consistency of MMFS-GA for feature selection. IMTFS obtained the best average balanced accuracy in both the standard and repeated datasets, 92.7\%, among the five competing multi-view feature selection methods, indicating that inter-view information preservation is important for feature selection. \revised{Detailed results from each of the ten experiments, including individual experiment accuracies, have been provided in Section 2.1 of the supplementary material Table 3.}
\begin{table*}[!ht]

\footnotesize
\begin{center}
\caption{Comparison of the MMFS-GA algorithm with various baseline approaches for binary classification on multi-view synthetic datasets. The table displays the average PCC and its standard deviation across 10 experiments. There were 100 training samples per class. 'Synthetic Data' represents the baseline scenario, while 'Synthetic Data (rep)' refers to experiments incorporating repeated informative views in the dataset.}

\setlength{\tabcolsep}{9.0pt} 
\renewcommand{\arraystretch}{1.7} 
\resizebox{\textwidth}{!}{\begin{tabular}{l c c c c c c }
\hline
\hline
\multirow{2}{*}{Experiments}   & \multicolumn{6}{c}{Accuracy(Binary)} \\ \cline{2-7}  
  & M2TFS & IMTFS & \makecell{LASSO-SVM} & ASMFS & SFFS & MMFS-GA\\

 \hline
Synthetic Data   & 0.89 $\pm$ 0.022 & 0.92 $\pm$ 0.007 & 0.91 $\pm$ 0.014 & 0.90 $\pm$ 0.015 & 0.91 $\pm$ 0.031 & $\textbf{0.96}$ $\pm$ 0.003 \\

Synthetic Data (rep)  &  0.87 $\pm$ 0.025 & 0.92 $\pm$ 0.008 & 0.92 $\pm$ 0.012 & 0.87 $\pm$ 0.017  & 0.86 $\pm$ 0.042 & \textbf{0.95} $\pm$ 0.011 \\
\hline

\hline

\end{tabular}}
\label{Tab_4:Binary_BalAcc_SynData}
\end{center}
\vspace{0mm}
\end{table*}

\begin{figure*}[h]
\centering
\includegraphics[width=0.99\textwidth]{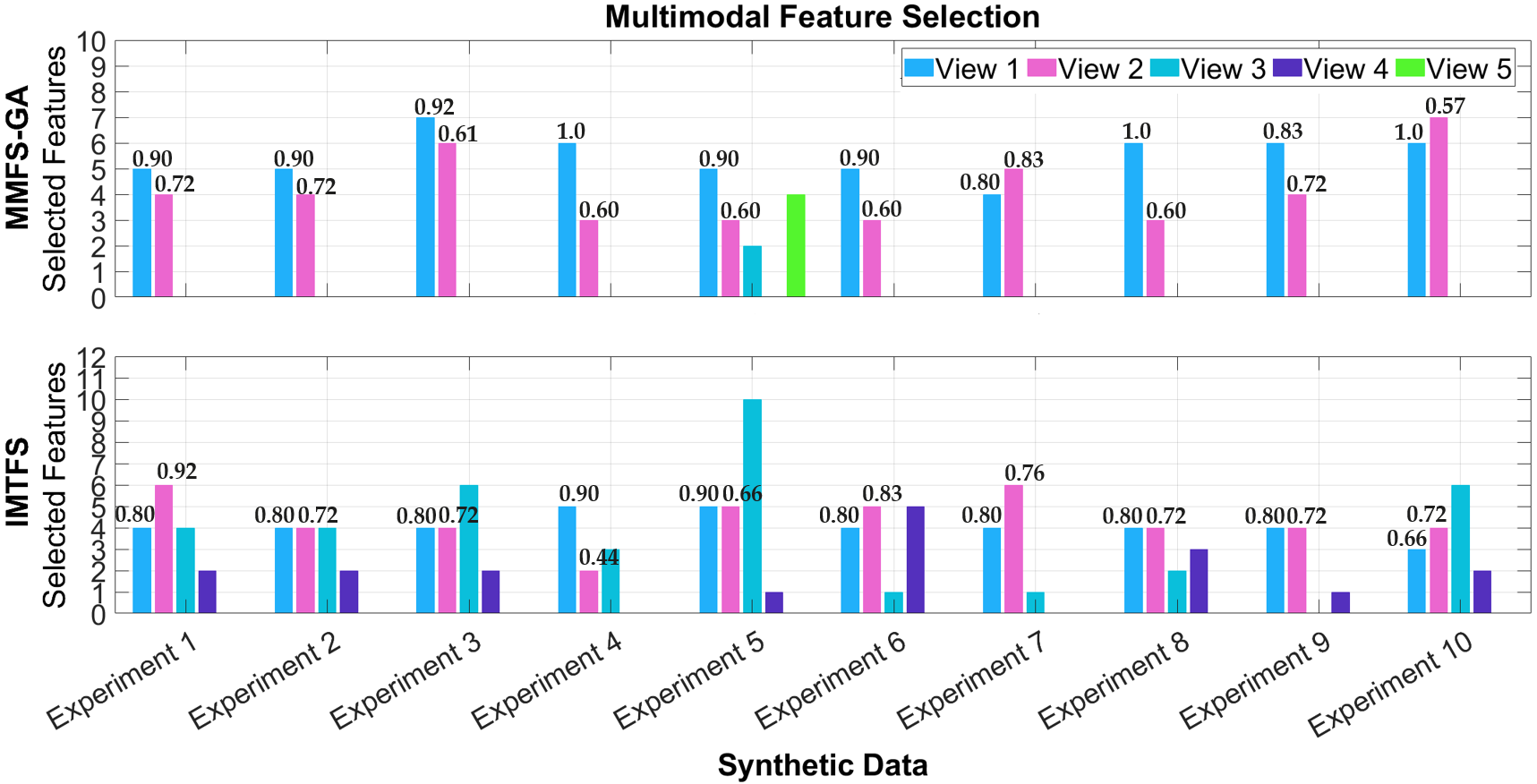}
\caption{Comparison of the MMFS-GA algorithm against the best baseline methods for binary classification. The F1 scores between informative and selected features are shown at the top of each bar. Blue bars represent the number of selected features from the view with high discriminative power; pink bars represent the number of selected features from the view with low discriminative power; and cyan, violet, green indicate selections from noise groups.}
\label{Fig_4:F1_Binary_SynData}
\vspace{0mm}
\end{figure*}

Figure~\ref{Fig_4:F1_Binary_SynData} displays the number of selected features from each feature view, together with a comparison of the F1 scores between the actual features and the selected features in the views with discriminative power (Views 1 and 2). Figure~\ref{Fig_4:F1_Binary_SynData} illustrates that MMFS-GA was more effective than the best competing method (i.e. IMTFS) in exploiting the relationships between different views. MMFS-GA was successful in filtering out views of pure noise features that were not relevant to the final classification task in nine out of ten experiments, indicating its effectiveness in distinguishing and removing irrelevant feature views. In contrast, the other methods assigned non-zero weights to the features in irrelevant feature groups. For example, with the IMTFS method, high weights have been assigned to the group of random uniform and Chi-square noise. Furthermore, IMTFS, M2TFS, and ASMFS all require the same number of features for each view, which might restrict their application. 

\subsubsection{Synthetic Dataset with Views of Different Dimensionality}

We also evaluated the MMFS-GA algorithm in the case where views have different numbers of features\revised{, thereby simulating more realistic and varied datasets}. Since the competing multi-view feature selection comparison techniques require the same number of features from each view, we compared our algorithm with only the Lasso-SVM algorithm. To generate multi-view data with different feature sizes, we intentionally introduced noise into some of the views and removed noise from others. This allowed us to simulate realistic scenarios where each view of the data has a distinct set of features and to evaluate the robustness of our methods to variations in feature size and quality. We report the classification accuracy and the F1 score between the informative and selected features in Table~\ref{Tab_5:Binary_Diff_SynData}. The results demonstrate that MMFS-GA outperformed Lasso-SVM in all experiments. Our proposed algorithm also exhibited high stability in selecting informative features compared to Lasso-SVM, demonstrating its robustness to variations in the number of features per view. Detailed results from each of the ten experiments, including individual experiment accuracies, have been provided in Section 2.3 of the supplementary material in Table 5. 
\begin{table*}[!ht]
\footnotesize
\begin{center}
\caption{Classification performance on synthetic multi-view data with distinct dimensions for each view across 10 experiments. For the first two views, the mean and standard deviation of the $F1$-scores between informative features and selected features are displayed. SF denotes the median of the total number of selected features in each view.}
\setlength{\tabcolsep}{10.0pt} 
\renewcommand{\arraystretch}{1.5} 
\resizebox{\textwidth}{!}{\begin{tabular}{l c c c c c c c c}
\hline
\hline
\multirow{3}{*}{Method}& \multicolumn{7}{c}{Group views} &\multirow{3}{*}{Accuracy}\\ \cline{2-8}  
& \multicolumn{2}{c}{View 1} & \multicolumn{2}{c}{View 2} & View 3 & View 4 & View 5 & \\\cline{2-8}
& $F1$ & SF & $F1$ & SF & SF & SF & SF &\\
 \hline

MMFS-GA & \makecell{\textbf{0.88} $\pm$\\ 0.05} & 5 & \makecell{\textbf{0.65} $\pm$\\ 0.10} & 5 & 0 & 0 & 0 & \makecell{\textbf{0.96} $\pm$\\ 0.00} \\ 
LASSO-SVM & \makecell{0.77 $\pm$\\ 0.05} & 4 & \makecell{0.46 $\pm$\\ 0.16} & 3 & 0 & 0 & 0 & \makecell{0.91 $\pm$\\ 0.01} \\

\hline

\hline

\hline

\end{tabular}}
\label{Tab_5:Binary_Diff_SynData}
\end{center}
\vspace{0mm}
\end{table*}

\subsubsection{TCGA Dataset binary}

The MMFS-GA algorithm's binary classification results for TCGA cancer types are presented in Table~\ref{Tab_5:TCGA}. 
In the BRCA dataset, the MMFS-GA algorithm demonstrated outstanding performance with a balanced accuracy of 99.6\%, tying with ASMFS for the highest accuracy and surpassing other methods. This result highlights the algorithm's precision in identifying breast cancer markers, a critical area in oncological diagnostics. For the LGG dataset, representing lower grade gliomas, MMFS-GA achieved the highest accuracy of 98.9\%. This performance, notably higher than the other methods, underscores the capability of MMFS-GA in discerning subtle yet critical genomic differences, which is paramount in the classification of glioma subtypes. In the PRAD dataset, focused on prostate adenocarcinoma, MMFS-GA again showed top-tier performance, matching the highest accuracy of 98.7\% achieved by M2TFS. This suggests the algorithm's robustness in handling different cancer types, especially those with complex genetic backgrounds. The analysis on the THCA dataset, pertaining to thyroid carcinoma, further established MMFS-GA as a leading method with an accuracy of 98.9\%, the highest among the compared methods. This indicates the algorithm’s effectiveness in distinguishing thyroid cancer, which often presents diagnostic challenges due to its diverse histological features. Across all datasets, MMFS-GA consistently ranked at the top in terms of accuracy, demonstrating its versatility and reliability in binary classification tasks. 
\begin{table*}[!ht]

\footnotesize
\begin{center}
\caption{Comparison of the MMFS-GA algorithm with various baseline approaches for binary classification on TCGA datasets. }

\setlength{\tabcolsep}{9.0pt} 
\renewcommand{\arraystretch}{1.7} 
\resizebox{\textwidth}{!}{\begin{tabular}{l c c c c c c }
\hline
\hline
\multirow{2}{*}{Experiments}   & \multicolumn{6}{c}{Accuracy(Binary)} \\ \cline{2-7}  
  & M2TFS & IMTFS & \makecell{LASSO-SVM} & ASMFS & SFFS & MMFS-GA\\

 \hline
BRCA   & 0.994 $\pm$ 0.01 &   0.994 $\pm$ 0.01  &   0.988 $\pm$ 0.02  &   \textbf{0.996} $\pm$ 0.01   &  0.975 $\pm$ 0.01   & \textbf{0.996} $\pm$ 0.01  \\

LGG   & 0.976 $\pm$ 0.02 &   0.969 $\pm$ 0.02  & 0.978 $\pm$ 0.02 &  0.982 $\pm$ 0.02 & 0.977 $\pm$ 0.02 & \textbf{0.989} $\pm$ 0.00 \\

PRAD   & \textbf{0.987} $\pm$ 0.011 & 0.984 $\pm$ 0.015 & 0.982 $\pm$ 0.016 & 0.985 $\pm$ 0.011 & 0.980 $\pm$ 0.017 & \textbf{0.987} $\pm$ 0.02 \\

THCA   & 0.967 $\pm$ 0.04 & 0.975 $\pm$ 0.03 & 0.984 $\pm$ 0.02 & 0.979 $\pm$ 0.03 & 0.969 $\pm$ 0.04 & \textbf{0.989} $\pm$ 0.00 \\
\hline

\hline

\end{tabular}}
\label{Tab_5:TCGA}
\end{center}
\vspace{0mm}
\end{table*}
\subsubsection{TADPOLE Dataset}
 
\begin{figure*}[h!]
\centering
\includegraphics[width=0.99\textwidth]{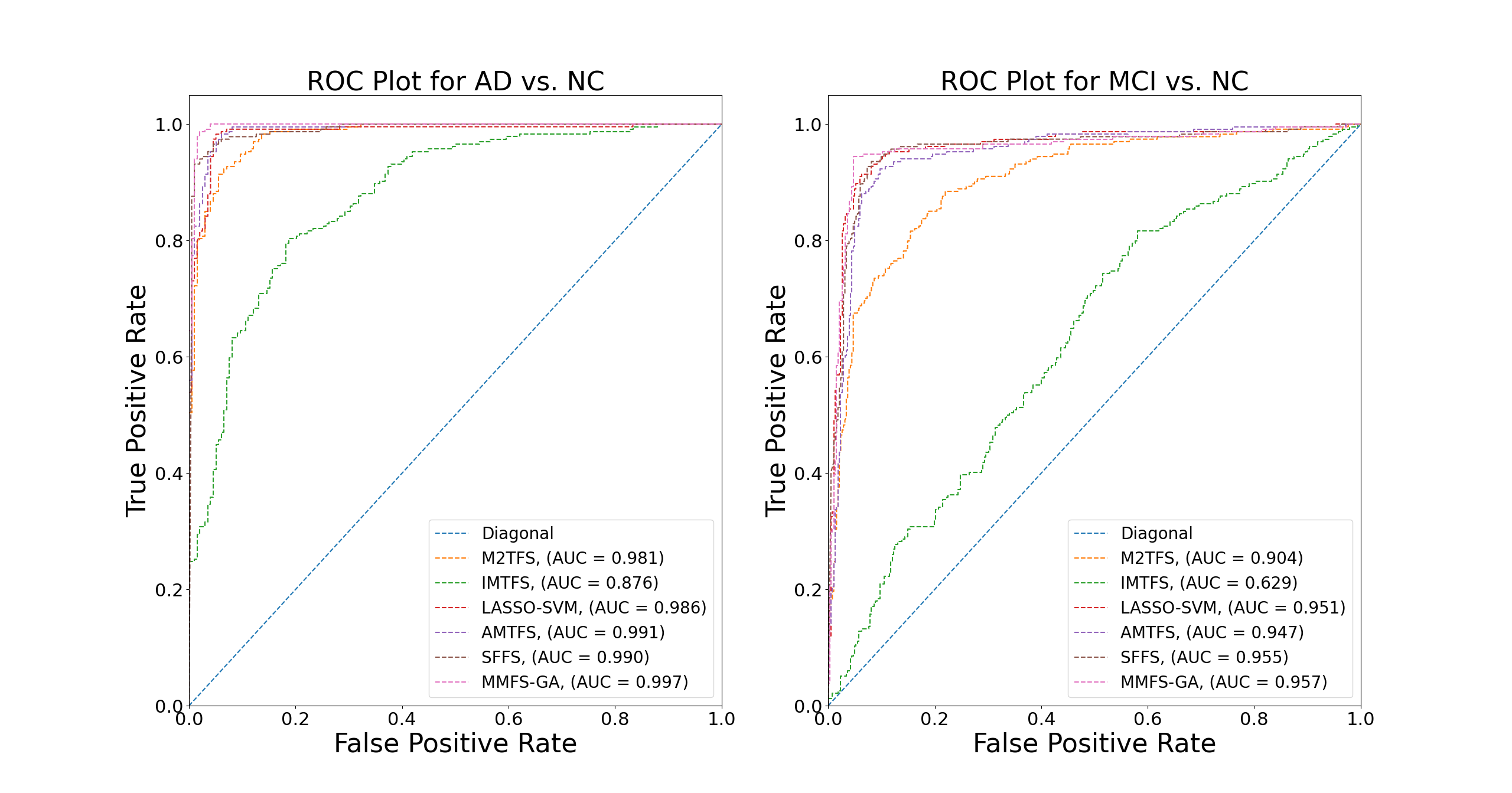}
\caption{ROC curves of all classification methods for AD versus NC and MCI versus NC. Compared to existing approaches, the proposed method obtains the best performance with the highest AUC, a high true positive rate (TPR), and a low false positive rate (FPR).} 
\label{Fig_5:ROC_Bin}
\vspace{0mm}
\end{figure*}

The detailed classification results in the TADPOLE dataset are summarized in Table~\ref{Tab_6:Binary_TADPOLE}. In particular, for the AD versus NC classification, the proposed method achieves a balanced accuracy of 97.9\%, a sensitivity of 98.3\%, a specificity of 97.5\%, and an area under the ROC curve (AUC) of 0.997, demonstrating excellent diagnostic efficacy. In comparison, our method has the highest sensitivity and specificity, demonstrating that it rarely overlooks an AD patient or incorrectly labels a healthy person as diseased. Moreover, the proposed method for the classification of MCI from NC achieved a balanced accuracy of 94.9\%, a sensitivity of 92.5\%, a specificity of 96.6\%, and an AUC of 0.957, while the best performance in comparative methods is 92.5\% by SVM-Lasso and then 90.7\% by ASMFS. Figure~\ref{Fig_5:ROC_Bin} plots the corresponding Receiver Operating Characteristic (ROC) curves, reflecting the superiority of our method compared to the other five methods. 
\begin{table*}[!ht]
\footnotesize
\begin{center}
\caption{Comparison of the MMFS-GA algorithm with baseline methods for AD vs. NC and MCI vs. NC classification on TADPOLE dataset based on balanced accuracy, sensitivity (SEN), specificity (SPE), and area under the curve (AUC). The baseline levels, predicting the 36 month diagnosis based on 24 month diagnoses, for the first and second experiments are set at 96.7\% and 94.8\%, respectively.}
\setlength{\tabcolsep}{18.5pt} 
\renewcommand{\arraystretch}{1.1} 
\resizebox{\textwidth}{!}{\begin{tabular}{l c c c c }
\hline
\hline
AD versus NC & & & &  \\ \cline{1-5}
Method   & \makecell{Balanced \\ Accuracy} & SEN & SPE & AUC \\ \cline{1-5}    
 \hline
M2TFS & \makecell{0.921\\(0.896 to 0.947)} & \makecell{0.949\\(0.926 to 0.968)} & \makecell{0.894\\(0.863 to 0.921)} & \makecell{0.981\\(0.968 to 0.993)} \\

IMTFS & \makecell{0.797\\(0.759 to 0.833)} & \makecell{0.816\\(0.780 to 0.852)} & \makecell{0.778\\(0.738 to 0.817)} & \makecell{0.876\\(0.845 to 0.907)} \\

LASSO-SVM & \makecell{0.966\\(0.949 to 0.981)} & \makecell{\textbf{0.983}\\(0.970 to 0.993)} & \makecell{0.949\\(0.928 to 0.970)} & \makecell{0.986\\(0.975 to 0.995)} \\

ASMFS & \makecell{0.957\\(0.938 to 0.975)} & \makecell{0.974\\(0.958 to 0.988)} & \makecell{0.939\\(0.917 to 0.961)} & \makecell{0.991\\(0.981 to 0.998)} \\

SFFS & \makecell{0.957\\(0.938 to 0.975)} & \makecell{0.944\\(0.921 to 0.965)} & \makecell{0.970\\(0.954 to 0.984)} & \makecell{0.990\\(0.979 to 0.998)} \\

MMFS-GA & \makecell{\textbf{0.979}\\(0.965 to 0.991)} & \makecell{0.979\\(0.963 to 0.991)} & \makecell{\textbf{0.980}\\(0.965 to 0.991)} & \makecell{\textbf{0.997}\\(0.991 to 1.000)} \\

\hline

\hline
 
MCI versus NC & & & &  \\ \cline{1-5}
Method   & \makecell{Balanced \\ Accuracy} & SEN & SPE & AUC \\ \cline{1-5}  
  
 \hline
M2TFS & \makecell{0.820\\(0.789 to 0.851)} & \makecell{0.782\\(0.748 to 0.815)} & \makecell{0.859\\(0.831 to 0.886)} & \makecell{0.904\\(0.880 to 0.927)} \\

IMTFS & \makecell{0.579\\(0.541 to 0.619)} & \makecell{0.543\\(0.503 to 0.583)} & \makecell{0.615\\(0.576 to 0.654)} & \makecell{0.629\\(0.591 to 0.667)} \\

LASSO-SVM & \makecell{0.925\\(0.904 to 0.945)} & \makecell{0.910\\(0.886 to 0.932)} & \makecell{0.940\\(0.920 to 0.958)} & \makecell{0.951\\(0.933 to 0.968)} \\

ASMFS & \makecell{0.907\\(0.883 to 0.930)} & \makecell{0.906\\(0.883 to 0.929)} & \makecell{0.908\\(0.885 to 0.930)} & \makecell{0.947\\(0.929 to 0.964)} \\

SFFS & \makecell{0.918\\(0.896 to 0.938)} & \makecell{0.906\\(0.883 to 0.929)} & \makecell{0.929\\(0.907 to 0.948)} & \makecell{0.955\\(0.938 to 0.971)} \\

MMFS-GA & \makecell{0.949\\(0.930 to 0.966)} & \makecell{0.944\\(0.925 to 0.961)} & \makecell{0.953\\(0.935 to 0.969)} & \makecell{0.957\\(0.940 to 0.972)} \\

\hline

\hline
\end{tabular}}
\label{Tab_6:Binary_TADPOLE}
\end{center}
\vspace{0mm}
\end{table*}

\subsubsection{Prediction of Conversion from MCI to Dementia}

\begin{figure*}[!h]
\centering
\includegraphics[width=0.99\textwidth]{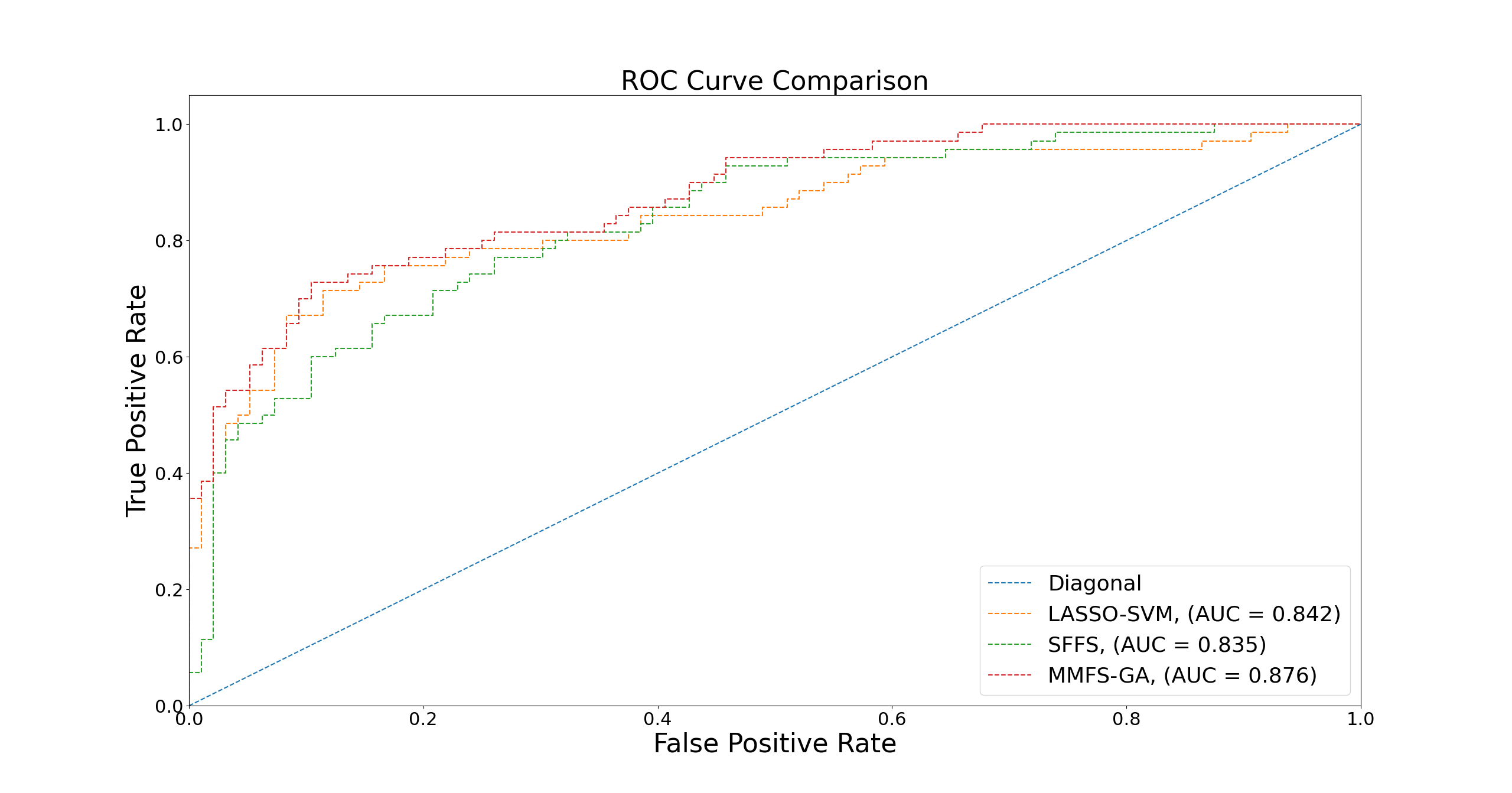}
\caption{ROC curves of the MMFS-GA algorithm and baseline methods for predicting conversion from MCI to dementia. The ROC curves were generated by plotting the true positive rate against the false positive rate for each method, based on the predicted probabilities. The AUC values are also shown in the legend.} 
\label{Fig_6:ROC_MCI_Conversion}
\vspace{0mm}
\end{figure*}

We evaluated the performance of our proposed algorithm, MMFS-GA, for predicting the conversion from MCI to dementia and compared it with two baseline approaches, LASSO-SVM and SFFS. Table~\ref{Tab_7:Binary_Conversion} presents the comparison of the three methods based on balanced accuracy, sensitivity, specificity, and AUC. Our MMFS-GA algorithm achieved the highest balanced accuracy of 0.805, which was higher than the values obtained by LASSO-SVM (0.784) and SFFS (0.745). Similarly, the sensitivity and specificity of our method were 0.714 and 0.896, respectively, which were higher than those of LASSO-SVM (0.671 and 0.896) and SFFS (0.657 and 0.833). The AUC value of our MMFS-GA algorithm was the highest among the three methods, with a value of 0.876, indicating its superior performance in distinguishing between MCI patients who will and will not convert to dementia. Furthermore, we visualized the performance comparison using ROC curves. As shown in Figure~\ref{Fig_6:ROC_MCI_Conversion}, the ROC curve of our MMFS-GA algorithm was consistently higher and closer to the top-left corner than those of the baseline methods, indicating its superior discriminative ability in predicting the conversion from MCI to dementia.
\begin{table*}[!ht]
\footnotesize
\begin{center}
\caption{Comparison of the MMFS-GA algorithm with baseline approaches for predicting conversion from MCI to dementia. The performance of each method was evaluated based on balanced accuracy, sensitivity (SEN), specificity (SPE), and area under the curve (AUC).}

\setlength{\tabcolsep}{18.5pt} 
\renewcommand{\arraystretch}{1.2} 
\resizebox{\textwidth}{!}{\begin{tabular}{l c c c c }
\hline
Method   & \makecell{Balanced \\ Accuracy} & SEN & SPE & AUC \\ \cline{1-5}  
  
 \hline

LASSO-SVM & \makecell{0.784\\(0.752 to 0.817)} & \makecell{0.671\\(0.633 to 0.708)} & \makecell{\textbf{0.896}\\(0.872 to 0.919)} & \makecell{0.842\\(0.813 to 0.870)} \\

SFFS & \makecell{0.745\\(0.709 to 0.779)} & \makecell{0.657\\(0.619 to 0.695)} & \makecell{0.833\\(0.804 to 0.862)} & \makecell{0.835\\(0.805 to 0.864)} \\

MMFS-GA & \makecell{\textbf{0.805}\\(0.773 to 0.836)} & \makecell{\textbf{0.714}\\(0.679 to 0.750)} & \makecell{\textbf{0.896}\\(0.872 to 0.919)} & \makecell{\textbf{0.876}\\(0.849 to 0.901)} \\                                  
\hline

\hline
\end{tabular}}
\label{Tab_7:Binary_Conversion}
\end{center}
\vspace{0mm}
\end{table*}

\subsection{Multiclass Classification Results}
\subsubsection{Synthetic Dataset}
Table~\ref{Tab_8:Multiclass_BalAcc_SynData}, lists the balanced accuracy of our method compared to the competing methods in the 4-class classification task. The overall estimate of the CER was calculated for each method based on the 10 test sets evaluated. The results for M2TFS, IMTFS, LASSO-SVM, LASSO-MkSVM, SFFS, and MMFS-GA were 0.32, 0.30, 0.16, 0.24, 0.50, and 0.10, respectively. MMFS-GA achieved the lowest CER among all methods, indicating superior performance in terms of classification accuracy. These results demonstrate the effectiveness of our proposed method, MMFS-GA, in handling feature selection and classification tasks in high-dimensional datasets. The results and accuracies of the ten experiments are detailed in Section 2.2 of the supplementary material Table 4.

\begin{table*}[!ht]
\footnotesize
\begin{center}
\caption{Comparison of the MMFS-GA algorithm with baseline approaches for 4-class classification using multi-view synthetic datasets. The table displays the average PCC and its standard deviation across 10 experiments. There were 100 training samples per class. 'Synthetic Data' represents the baseline scenario, while 'Synthetic Data (rep)' refers to experiments incorporating repeated informative views in the dataset. As a four-class classification problem was analyzed, the accuracy for random guessing would be 25\%.}

\setlength{\tabcolsep}{15.0pt} 
\renewcommand{\arraystretch}{1.5} 
\resizebox{\textwidth}{!}{\begin{tabular}{l c c c c c c }
\hline
\hline
\multirow{3}{*}{Experiments}   & \multicolumn{6}{c}{Accuracy(4-class)} \\ \cline{2-7}  
  & M2TFS & IMTFS & \makecell{LASSO-SVM} & \makecell{LASSO-MkSVM} & SFFS & MMFS-GA\\
  \hline

Synthetic Data  & 0.64 $\pm$ 0.006  & 0.74 $\pm$ 0.008  & 0.82 $\pm$ 0.024 & 0.76 $\pm$ 0.030 & 0.68 $\pm$ 0.052 & $\textbf{0.90}$ $\pm$ 0.022\\
Synthetic Data (rep)   &  0.52 $\pm$ 0.103 & 0.73 $\pm$ 0.183 & 0.78 $\pm$ 0.018 & 0.74 $\pm$ 0.010 & 0.68 $\pm$ 0.052  & \textbf{0.91} $\pm$ 0.014\\

\hline
\hline

\hline

\end{tabular}}
\label{Tab_8:Multiclass_BalAcc_SynData}
\end{center}
\vspace{0mm}
\end{table*}

\begin{figure*}[h!]
\centering
\includegraphics[width=0.99\textwidth]{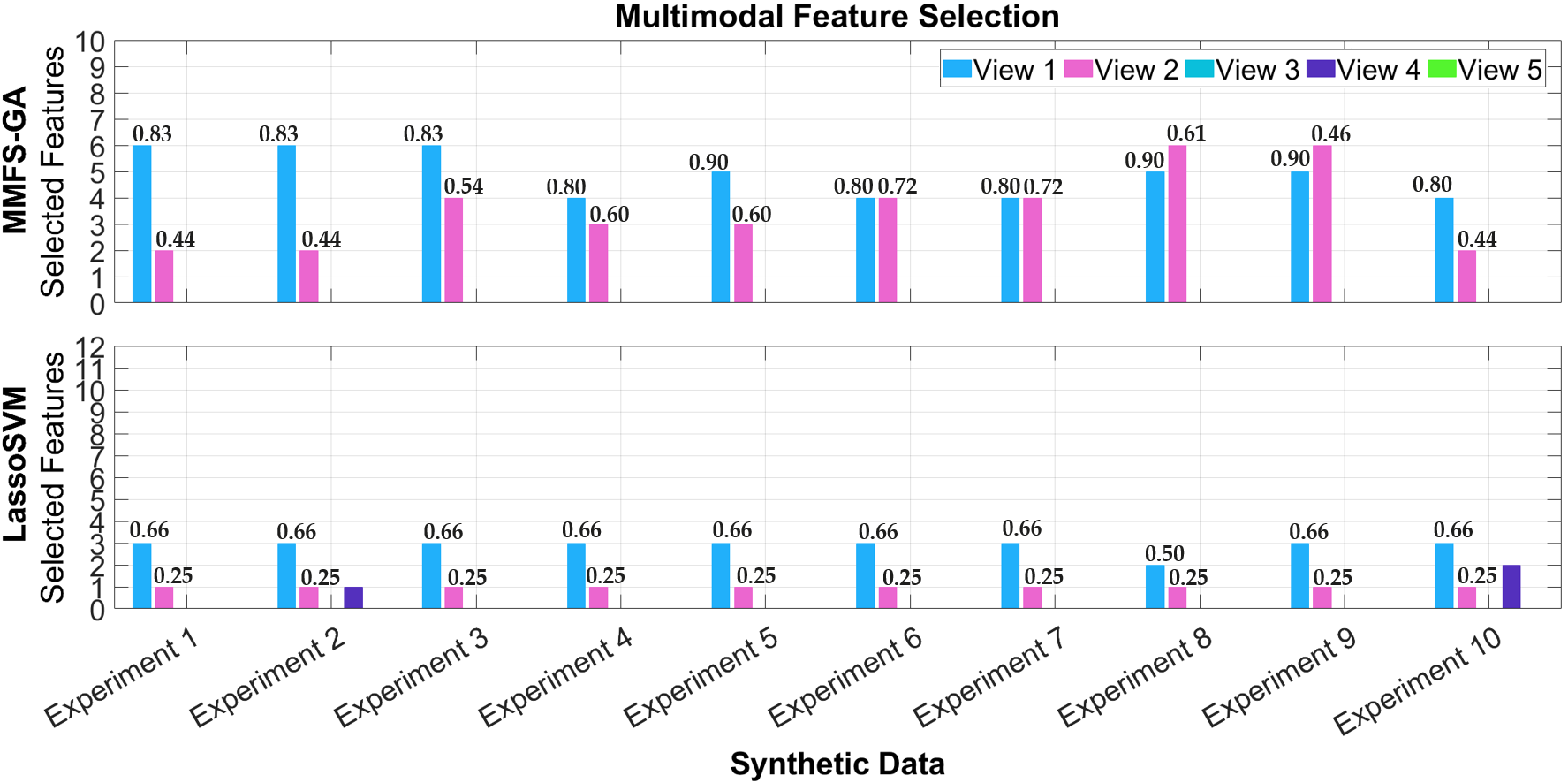}
\caption{Comparison of selected features between the MMFS-GA algorithm and Lasso-SVM as the best baseline method for 4-class classification utilizing ten experiments of synthetic data. The F1 scores between informative features and selected features are shown at the top of each bar. The blue bars represent the number of selected features from the feature group with high discriminative power and the pink bars represent the number of selected features from the feature group with low discriminative power. The cyan, violet, and green bars represent the number of features selected from the noisy feature groups.} 
\label{Fig_7:F1_4Class_SynData}
\vspace{0mm}
\end{figure*}

Figure~\ref{Fig_7:F1_4Class_SynData} presents a comparison of F1 scores between the actual informative features and those that have been selected for MMFS-GA and the best performing competing method. As the F1 scores in Figure~\ref{Fig_7:F1_4Class_SynData} indicate, MMFS-GA was more effective than Lasso-SVM in filtering the noisy features from the feature groups with discriminative power and exploiting the relationships among different views. More specifically, the average F1 scores for our proposed MMFS-GA method were 0.84 and 0.56 for views 1 and 2, respectively, while the average F1 scores for the Lasso-SVM method were 0.64 and 0.25 for views 1 and 2, respectively. 

\subsubsection{Synthetic Dataset with Views of Different Dimensionality}
We evaluated the MMFS-GA algorithm in the case that views have different numbers of features, similarly to the experiments we conducted for binary classification. We report the classification accuracy and the F1 score between the informative and selected features in Table~\ref{Tab_9:DIF_Mul_SynData}.  Our results indicate that MMFS-GA outperformed Lasso-SVM in terms of both. Specifically, the average accuracy of MMFS-GA was 0.86, whereas the average accuracy of Lasso-SVM was 0.82. This suggests that MMFS-GA is a more effective method for adapting to varying feature sizes in multiclass classification problems. Furthermore, we compared the F1 score for informative and selected features between MMFS-GA and Lasso-SVM in View 1 and View 2. The average F1 score in View 1 was 0.83 in MMFS-GA, whereas it was reduced to 0.63 in Lasso-SVM. Similarly, in View 2, the average F1 score was 0.58 in MMFS-GA, while Lasso-SVM achieved an average F1 score of 0.25. These results suggest that the MMFS-GA algorithm is more effective at selecting informative features than the Lasso-SVM algorithm, which may lead to better classification performance. The results and accuracies of the ten experiments are detailed in Section 2.4 of the supplementary material Table 6.
\begin{table*}[!ht]
\footnotesize
\begin{center}
\caption{Classification performance on synthetic multi-view data  with distinct dimensions for each view using LASSO-SVM. For the first two views, the F1 scores between informative features and selected features are displayed. SF denotes the total number of selected features in each view.}

\setlength{\tabcolsep}{10.0pt} 
\renewcommand{\arraystretch}{1.5} 
\resizebox{\textwidth}{!}{\begin{tabular}{l c c c c c c c c}
\hline
\hline
\multirow{3}{*}{Method}& \multicolumn{7}{c}{Group views} &\multirow{3}{*}{Accuracy}\\ \cline{2-8}  
& \multicolumn{2}{c}{View 1} & \multicolumn{2}{c}{View 2} & View 3 & View 4 & View 5 & \\\cline{2-8}
& $F1$ & SF & $F1$ & SF & SF & SF & SF &\\
 \hline

MMFS-GA & \makecell{\textbf{0.83} $\pm$\\ 0.06} & 4 & \makecell{\textbf{0.58} $\pm$\\ 0.10} & 4 & 0 & 0 & 0 & \makecell{\textbf{0.86} $\pm$\\ 0.01} \\ 
LASSO-SVM & \makecell{0.65 $\pm$\\ 0.05} & 3 & \makecell{0.25 $\pm$\\ 0.00} & 1 & 0 & 0 & 0 & \makecell{0.82 $\pm$\\ 0.01} \\

\hline

\hline

\end{tabular}}
\label{Tab_9:DIF_Mul_SynData}
\end{center}
\vspace{0mm}
\end{table*}
\subsubsection{TCGA Dataset}
We evaluated the MMFS-GA algorithm's performance on multiclass classification using various TCGA cancer datasets (BRCA, LGG, OVCA, and RCC), as detailed in Section~\ref{TCGA}. 
The results, detailed in Table~\ref{Tab_4:TCGA}, explain the comparative accuracy of MMFS-GA against several baseline methods, demonstrating its effective performance. In the BRCA dataset, MMFS-GA achieved an accuracy of 0.822, higher than the next best method, LASSO-MkSVM (0.736). In the LGG dataset, MMFS-GA's accuracy was high at 0.993, outperforming M2TFS (0.971). This highlights MMFS-GA's effectiveness in handling the heterogeneity of lower grade glioma. MMFS-GA's accuracy in the OVCA dataset was 0.867, better than LASSO-MkSVM's 0.755. This result reflects the algorithm's effectiveness in ovarian cancer classification. In the RCC dataset, MMFS-GA reached an accuracy of 0.998, compared to 0.956 for IMTFS, showing its high accuracy in renal cell carcinoma classification.

\begin{table*}[!ht]

\footnotesize
\begin{center}
\caption{Comparison of the MMFS-GA algorithm with various baseline approaches for multiclass classification on TCGA datasets. }

\setlength{\tabcolsep}{9.0pt} 
\renewcommand{\arraystretch}{1.7} 
\resizebox{\textwidth}{!}{\begin{tabular}{l c c c c c c }
\hline
\hline
\multirow{2}{*}{Experiments}   & \multicolumn{6}{c}{Accuracy(Multiclass)} \\ \cline{2-7}  
  & M2TFS & IMTFS & \makecell{LASSO-SVM} & LASSO-MkSVM & SFFS & MMFS-GA\\

 \hline
BRCA    & 0.716 $\pm$ 0.008 & 0.717 $\pm$ 0.012 & 0.541 $\pm$ 0.005  & 0.736 $\pm$ 0.006 & 0.575 $\pm$ 0.006 &  \textbf{0.822} $\pm$ 0.025  \\

LGG   & 0.971 $\pm$ 0.005 & 0.966 $\pm$ 0.004 & 0.948 $\pm$ 0.002 & 0.955 $\pm$ 0.002 & 0.810 $\pm$ 0.004 &  \textbf{0.993} $\pm$ 0.002  \\

OVCA   & 0.502 $\pm$ 0.018 & 0.507 $\pm$ 0.025 & 0.620 $\pm$ 0.010 & 0.755 $\pm$ 0.009 & 0.422 $\pm$ 0.009 &  \textbf{0.867} $\pm$ 0.016  \\

RCC   & 0.955 $\pm$ 0.005 & 0.956 $\pm$ 0.004 & 0.923 $\pm$ 0.002 & 0.928 $\pm$ 0.002 & 0.703 $\pm$ 0.002  &  \textbf{0.998} $\pm$ 0.001  \\

\hline

\hline

\end{tabular}}
\label{Tab_4:TCGA}
\end{center}
\vspace{0mm}
\end{table*}

\subsubsection{TADPOLE Dataset}
We employed the multi-view TADPOLE dataset in order to verify the proposed method for classifying between NC vs MCI vs AD as three class classification tasks. The results of the proposed MMFS-GA for multiclass classification with the TADPOLE dataset are shown in Table~\ref{Tab_10:Multiclass_TADPOLE}, where our proposal outperforms all state-of-the-art baseline algorithms.
\begin{table*}[!ht]

\footnotesize
\begin{center}
\caption{Comparison of the MMFS-GA algorithm with baseline approaches for multiclass classification in the TADPOLE dataset, using balanced accuracy, true positive fractions (TPF), and AUC. Values in parentheses indicate 95\% confidence intervals.
The baseline levels, predicting the 36 month diagnosis based on 24 month diagnoses, for this experiment are set at 87.4\%.}
\setlength{\tabcolsep}{15.5pt} 
\renewcommand{\arraystretch}{1.5} 
\resizebox{\textwidth}{!}{\begin{tabular}{l c c c c c }
\hline
\hline

Method   & \makecell{Balanced \\ Accuracy} & TPF$_{NC}$ & TPF$_{MCI}$ & TPF$_{AD}$ & AUC \\ \cline{1-5}  
  
 \hline
M2TFS & \makecell{0.507\\(0.472 to 0.542)} & \makecell{0.521\\(0.457 to 0.583)} & \makecell{0.534\\(0.484 to 0.583)} & \makecell{0.465\\(0.396 to 0.535)} & \makecell{0.706\\(0.675 to 0.738)} \\

IMTFS & \makecell{0.508\\(0.473 to 0.546)} & \makecell{0.521\\(0.456 to 0.582)} & \makecell{0.529\\(0.481 to 0.577)} & \makecell{0.475\\(0.406 to 0.545)} & \makecell{0.703\\(0.672 to 0.734)} \\

LASSO-SVM & \makecell{0.855\\(0.831 to 0.882)} & \makecell{0.915\\(0.878 to 0.948)} & \makecell{0.929\\(0.901 to 0.953)} & \makecell{0.722\\(0.657 to 0.781)} & \makecell{0.899\\(0.854 to 0.946)} \\

LASSO-MkSVM & \makecell{0.794\\(0.763 to 0.821)} & \makecell{0.885\\(0.841 to 0.923)} & \makecell{0.814\\(0.777 to 0.849)} & \makecell{0.682\\(0.619 to 0.746)} & \makecell{0.870\\(0.827 to 0.913)} \\

SFFS & \makecell{0.818\\(0.793 to 0.845)} & \makecell{0.902\\(0.862 to 0.937)} & \makecell{\textbf{0.940}\\(0.916 to 0.962)} & \makecell{0.611\\(0.544 to 0.675)} & \makecell{0.915\\(0.894 to 0.933)} \\

MMFS-GA   & \makecell{\textbf{0.881}\\(0.857 to 0.905)} & \makecell{\textbf{0.949}\\(0.919 to 0.976)} & \makecell{0.927\\(0.900 to 0.951)} & \makecell{\textbf{0.768}\\(0.706 to 0.822)} & \makecell{\textbf{0.962}\\(0.950 to 0.972)} \\

\hline

\hline

\end{tabular}}
\label{Tab_10:Multiclass_TADPOLE}
\end{center}
\vspace{0mm}
\end{table*}

\begin{figure*}[h!]
\centering
\includegraphics[width=0.99\textwidth]{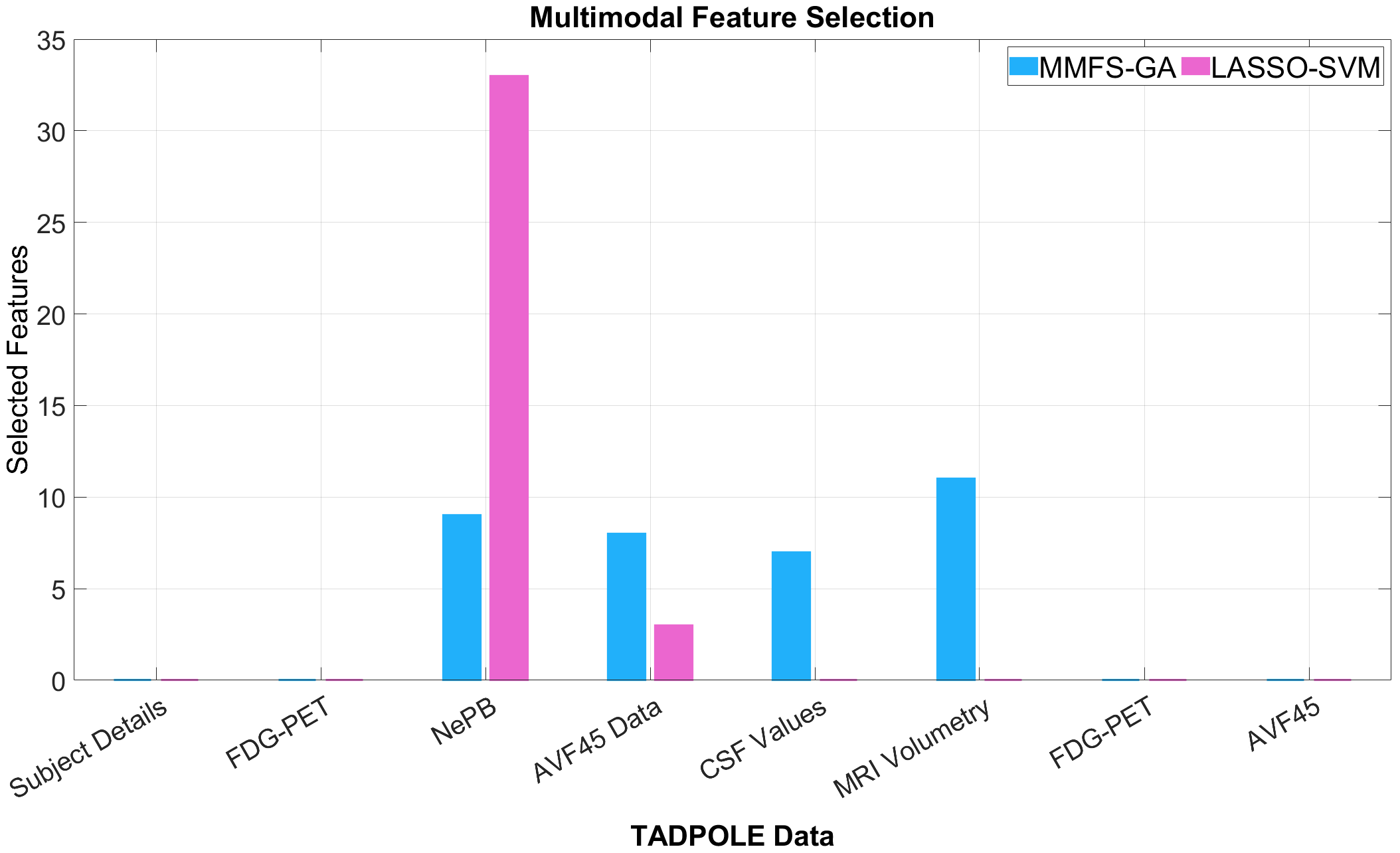}
\caption{Comparison of selected features between the MMFS-GA algorithm and Lasso-SVM as the best baseline method for multiclass classification utilizing the TADPOLE dataset.} 
\label{Fig_8:TADPOLE_Number_feature}
\vspace{0mm}
\end{figure*}

Figure~\ref{Fig_8:TADPOLE_Number_feature} presents the number of selected features from each feature view as well as the most informative combination of views. The ability of our proposed method to effectively use the global structural information included in multi-view data may be an essential reason for its superior performance in comparison to other baseline methods. In the MMFS-GA algorithm, the similarity of features between intra-views and between-views is limited by the objective function during evolution, which makes the selected features more informative and discriminative.

\section{Discussion}
\label{discussion}
The goal of multi-view machine learning algorithms is to create models that can jointly learn multiple views of data and relate information through a shared representation. Multi-view co-learning has been used effectively in several multi-disciplinary fields of machine learning, including computer vision~\citep{zhang2021natural}, speech recognition~\citep{muszynski2019recognizing}, and sub-fields of medical imaging~\citep{shi2022asmfs,shi2021c}, by allowing all views to influence the training and leveraging complementary information across views. However, sometimes multi-view data fusion comes with the curse of dimensionality, in which the number of features far exceeds the number of training subjects available, hence preventing the performance enhancement of multi-view learning algorithms~\citep{bellman2010dynamic}. Through this study, we intend to develop a novel multi-view feature selection algorithm for classification problems, one that not only eliminates irrelevant intra-view features, but also considers the inherent information of different data views and discards redundant data from between-views. Numerous studies~\citep{zu2016label,lei2017relational,yang2018incomplete,hao2020multi} have shown that the fusion of complementary information provided by a variety of data views can enhance classification performance. There are three fundamental limitations to these approaches, despite the fact that they provide good outcomes when working with multi-view data. This is because (1)  complementary information of different features of views is not adequately investigated, (2) statistical properties between-views are not considered, and (3) the performance of these methods will be reduced if one or more views of data are corrupted by noise.

In this study, we propose a multi-view multi-objective feature selection method based on a multi-chromosome genetic algorithm, MMFS-GA, to select discriminative features from each view by overcoming the above issues. The MMFS-GA method employs two multiniche multi-objective genetic algorithms to select the optimal set of views that yields the most stable solutions in terms of feature robustness and classification performance. The first GA selects the best solutions across all niches during the evolution period and at the end of the generation. In this way, we can observe the impact of combining features from other views of data on the classification performance, even those that were initially considered redundant. Once solutions have been generated for all views of the data, the second GA module looks for the best possible combination of solutions to obtain optimal classification performance. Since the second genetic algorithm benefits from a multiniche configuration, the optimal fusion solution can be determined across all niches according to the minimum number of features and the maximum classification accuracy. The proposed usage of MMFS-GA for the multi-view feature selection provides a framework capable of addressing various classification tasks, including binary and multiclass.

We evaluate the proposed MMFS-GA algorithm on both synthetic and real-world datasets using a simple multi-view classification problem with binary and multiclass tasks. We compare the performance of MMFS-GA with several baseline methods, including Adaptive-similarity-based multi-modality feature selection (ASMFS), Multi-kernel method with manifold regularized multitask feature learning (M2TFS), Multi-task learning-based feature selection by preserving inter-modality relationships (IMTFS), Multi-kernel method with Sparse Structure-Regularized Learning (Lasso-MkSVM), LASSO-based feature selection with multi-kernel SVM method (Lasso-SVM), and Sequential Forward Floating Search algorithm (SFFS). These baseline methods cover various aspects of feature selection and classification. Our experimental results, summarized in Tables \ref{Tab_4:Binary_BalAcc_SynData}--\ref{Tab_10:Multiclass_TADPOLE}, demonstrate that MMFS-GA outperforms all the baseline methods. MMFS-GA shows better performance in terms of noise reduction from intra-views and between-views, as well as selecting features with the highest discriminative power. Specifically, when comparing our method with Lasso-SVM in binary classification, we observed a slight difference in performance, with our method yielding slightly better results. However, in multiclass classification, our method consistently outperformed Lasso-SVM, demonstrating its potential for achieving improved results in multiclass classification tasks. Overall, our experiments suggest that the proposed MMFS-GA algorithm is a promising approach for multi-view feature selection and classification tasks on both synthetic and real-world datasets.

\section{Conclusion}
\label{conclusion}
In this paper, we have proposed a novel multi-view feature selection method, called the multi-view multi-objective feature selection genetic algorithm (MMFS-GA), which addresses the limitations of traditional feature selection methods for multi-view data for binary and multiclass classification tasks. The experimental results on synthetic and real-world datasets consistently demonstrate the remarkable performance compared to several existing state-of-the-art baseline methods. The MMFS-GA algorithm successfully reduces noise within and between views, and selects features with high discriminative power. The MMFS-GA algorithm utilizes two genetic algorithms with a parallel evolutionary structure, effectively and efficiently solving multi-view feature selection problems. In particular, the first GA selects effective features from each view, while the second explores complementary information between-views and intra-view aspects. In conclusion, this study highlights the effectiveness and efficiency of the MMFS-GA algorithm in addressing the challenges of multi-view feature selection, with the potential to enhance prediction models and improve generalization in various domains.

\section*{Acknowledgements}
This research has been supported by grants 316258, 346934, 332510, 358944 (Flagship of Advanced Mathematics for Sensing Imaging and Modelling) from the Research Council of Finland; grant 351849 from the Research Council of Finland under the frame of ERA PerMed ("Pattern-Cog"); Sigrid Juselius Foundation; the Jane and Aatos Erkko Foundation; grant 220104 from Jenny ja Antti Wihurin; grant 65221647 from Pohjois-Savon Rahasto; and the Doctoral Program in Molecular Medicine (DPMM) from the University of Eastern Finland.
The computational analyzes were performed on servers provided by UEF Bioinformatics Center, University of Eastern Finland, Finland.
Data collection and sharing for this project were funded by the Alzheimer's Disease Neuroimaging Initiative (ADNI) (National Institutes of Health Grant U01 AG024904) and DOD ADNI (Department of Defense award number W81XWH-12-2-0012). ADNI is funded by the National Institute on Aging, the National Institute of Biomedical Imaging and Bioengineering, and through generous contributions from the following: AbbVie, Alzheimer's Association; Alzheimer's Drug Discovery Foundation; Araclon Biotech; BioClinica, Inc.; Biogen; Bristol-Myers Squibb Company; CereSpir, Inc.; Cogstate; Eisai Inc.; Elan Pharmaceuticals, Inc.; Eli Lilly and Company; EuroImmun; F. Hoffmann-La Roche Ltd and its affiliated company Genentech, Inc.; Fujirebio; GE Healthcare; IXICO Ltd.; Janssen Alzheimer Immunotherapy Research \& Development, LLC.; Johnson \& Johnson Pharmaceutical Research \& Development LLC.; Lumosity; Lundbeck; Merck \& Co., Inc.; Meso Scale Diagnostics, LLC.; NeuroRx Research; Neurotrack Technologies; Novartis Pharmaceuticals Corporation; Pfizer Inc.; Piramal Imaging; Servier; Takeda Pharmaceutical Company; and Transition Therapeutics. The Canadian Institutes of Health Research is providing funds to support ADNI clinical sites in Canada. Private sector contributions are facilitated by the Foundation for the National Institutes of Health (www.fnih.org). The grantee organization is the Northern California Institute for Research and Education, and the study is coordinated by the Alzheimer's Therapeutic Research Institute at the University of Southern California. ADNI data are disseminated by the Laboratory for Neuro Imaging at the University of Southern California.
This work made use of the TADPOLE data sets \url{https://tadpole.grand-challenge.org} constructed by the EuroPOND consortium \url{http://europond.eu} funded by the European Union’s Horizon 2020 research and innovation program under grant agreement No 666992.

\section*{Data Availability}
The multimodal feature selection algorithms' source codes and the synthetic dataset are accessible at \url{https://github.com/vandadim/MMFS-GA}. Additionally, the Github repository contains the roster identification numbers (RIDs) of the subjects for the real dataset(\ref{Tadpole} and \ref{Conversion}) used in this study.

\end{document}